\documentclass[letterpaper]{article} 
\usepackage[submission]{aaai24}  
\usepackage{times}  
\usepackage{helvet}  
\usepackage{courier}  
\usepackage[hyphens]{url}  
\usepackage{graphicx} 
\usepackage{natbib}  
\usepackage{caption} 
\usepackage{xcolor}

\usepackage{ragged2e} 
\usepackage{booktabs,makecell, multirow, tabularx}
\usepackage{amsmath}
\usepackage{amsfonts}
\usepackage{subfigure}
\usepackage{bm}
\usepackage{floatrow}

\usepackage[label font=bf,labelformat=simple]{subfig}%

\usepackage{caption}

\usepackage{algorithm}
\usepackage{algorithmic}

\usepackage{newfloat}
\usepackage{listings}
\DeclareCaptionStyle{ruled}{labelfont=normalfont,labelsep=colon,strut=off} 
\floatstyle{ruled}
\newfloat{listing}{tb}{lst}{}
\floatname{listing}{Listing}
\title{Unleashing the Potential of Spiking Neural Networks for \\ Sequential Modeling with Contextual Embedding}
\author{
    Xinyi Chen$^1$,
    Jibin Wu$^1$\thanks{Corresponding Author: jibin.wu@polyu.edu.hk},
    Huajin Tang$^2$,
    Qinyuan Ren$^3$,
    Kay Chen Tan$^1$,
}
\affiliations{
    \textsuperscript{\rm 1}Department of Computing, The Hong Kong Polytechnic University\\
    \textsuperscript{\rm 2}College of Computer Science and Technology, Zhejiang University\\
    \textsuperscript{\rm 3}College of Control Science and Engineering, Zhejiang University\\

}

\usepackage{bibentry}

\begin{document}
\maketitle
\begin{abstract}
\begin{quote}
The human brain exhibits remarkable abilities in integrating temporally distant sensory inputs for decision-making. However, existing brain-inspired spiking neural networks (SNNs) have struggled to match their biological counterpart in modeling long-term temporal relationships. To address this problem, this paper presents a novel Contextual Embedding Leaky Integrate-and-Fire (CE-LIF) spiking neuron model. Specifically, the CE-LIF model incorporates a meticulously designed contextual embedding component into the adaptive neuronal firing threshold, thereby enhancing the memory storage of spiking neurons and facilitating effective sequential modeling. Additionally, theoretical analysis is provided to elucidate how the CE-LIF model enables long-term temporal credit assignment. Remarkably, when compared to state-of-the-art recurrent SNNs, feedforward SNNs comprising the proposed CE-LIF neurons demonstrate superior performance across extensive sequential modeling tasks in terms of classification accuracy, network convergence speed, and memory capacity. 
\end{quote}
\end{abstract}
\renewcommand{\thefootnote}{}
\footnotetext{Under review. Preprint}

\section{Introduction}

Driven by the recent advance in training algorithm development \cite{wu2018spatio,8891809} and model architecture design~\cite{zheng2021going,fang2021deep}, brain-inspired spiking neural networks (SNNs) have exhibited remarkable performance in tasks such as image classification~\cite{wu2021tandem,qin2023attention,wang2023adaptive}, speech processing~\cite{wu2018spiking,wu2020deep,wu2021progressive}, and robotic control~\cite{dewolf2021spiking,BDETT}. Furthermore, when implemented on ultra-low-power and densely-connected neuromorphic chips~\cite{merolla2014million,davies2018loihi,pei2019towards}, these models demonstrate compelling energy efficiency and low latency, making them highly promising for achieving ultra-low-power artificial intelligence (AI) at the edge.
However, despite these advancements, existing SNNs still face challenges in effectively modeling long-term temporal dependency when compared to their biological counterparts \cite{gutig2016spiking,qin2023attention}. 

Towards the goal of enhancing the sequential modeling capability of SNNs, recent research endeavors have focused on developing novel spiking neuron models that incorporate slow-decaying variables capable of storing long-term memory. One notable approach involves the integration of a threshold adaptation mechanism into spiking neurons, wherein the firing threshold of a neuron is elevated and gradually decays following each firing event~\cite{LSNN}. This adaptive firing threshold mechanism not only helps suppress excessive neuronal firing but also demonstrates effectiveness in storing long-term memory. Furthermore, Yin et al. propose to use a learnable decaying time constant for the adaptive firing threshold, thereby enabling the storage and utilization of multi-scale temporal information~\cite{ALIF}. Similarly, Shaban et al. introduce a novel adaptive spiking neuron model, wherein the threshold undergoes a double-exponential decay, enabling the storage of both short-term and long-term information~\cite{DEXAT}. However, despite the promising results observed with these models, their ability to establish long-term dependencies remains limited. This limitation can be attributed to the inadequate representation of the temporal structure of sensory inputs within the long-term memory created by the adaptive firing threshold. 

 \begin{figure*}[!ht]
\centering
\includegraphics[scale=0.4,trim= 0 6 0 5, clip]{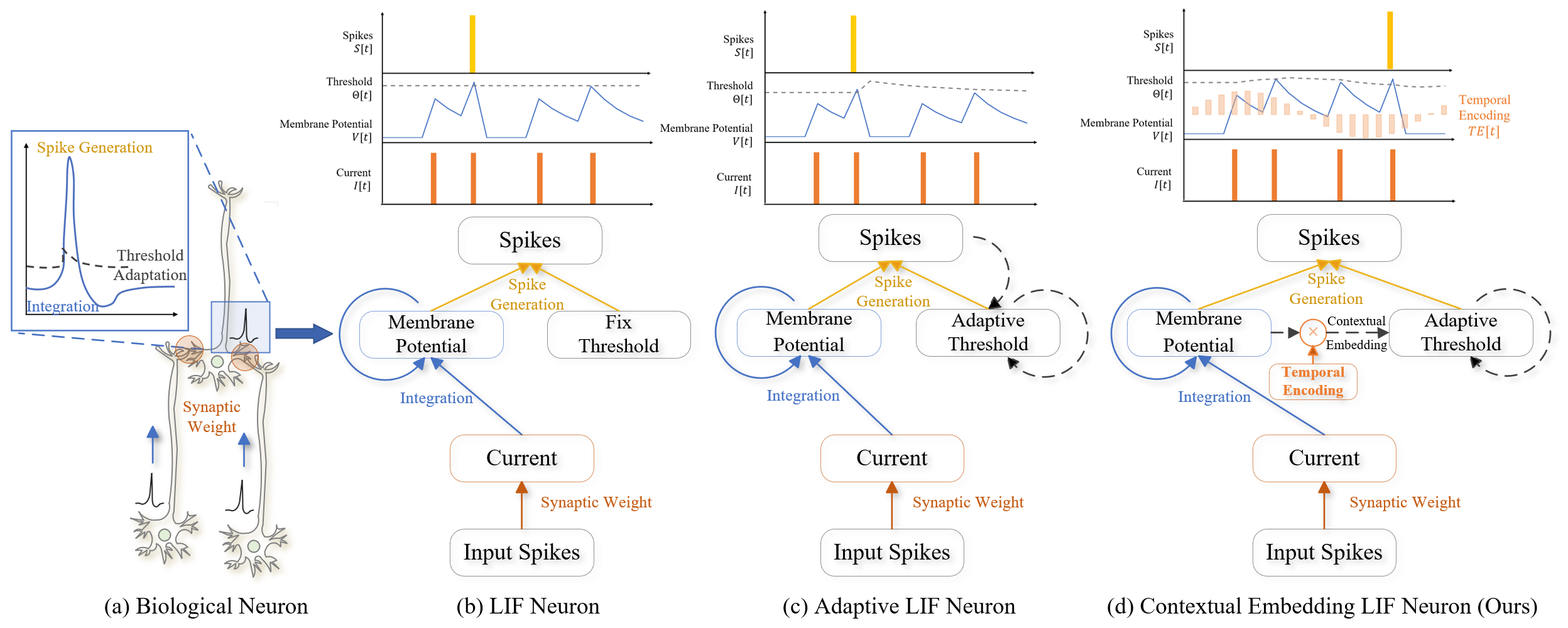}
\caption{The comparison of neuronal structure and dynamics between biological neurons and spiking neurons. \textbf{(a)} Biological neurons transduce presynaptic spikes into input currents, which are subsequently integrated into the membrane potential. The spike generation happens once the membrane potential surpasses the firing threshold. Notably, the firing threshold of biological neurons undergoes adaptations in response to the firing activities of the neuron. \textbf{(b-d)} The functional block diagrams (bottom row) and neuronal dynamics (top row) of the LIF neuron, ALIF neuron, and our proposed CE-LIF neuron. The LIF neuron generates spikes at a fixed firing threshold. In contrast, the ALIF model refines the LIF model by incorporating a biologically inspired adaptive firing threshold, thereby enhancing neuronal memory capacity while suppressing excessive firing. Furthermore, our proposed CE-LIF model advances upon the ALIF model by replacing the postsynaptic-spike-induced firing threshold adjustment with a contextual-embedding-induced one. Specifically, this contextual embedding is produced by multiplying the membrane potential with a learnable temporal encoding. This refinement not only results in enhanced memory capacity compared to ALIF but also facilitates long-term temporal credit assignment. 
}
\label{fig: main}
\end{figure*}

In another vein of research, the integration of self-attention mechanisms along with position encoding (PE) has been proposed in Transformer~\cite{vaswani2017attention} to enhance the efficacy of sequential modeling. Specifically, self-attention enables the direct modeling of temporal dependencies between distant inputs. Additionally, PE has been introduced to explicitly represent the temporal structure of inputs using a fixed-length vector, thereby addressing the position invariance issue associated with self-attention operations. In practice, PE can be generated through a fixed sinusoidal function~\cite{vaswani2017attention, GPT2}, or learned together with other model parameters \cite{gehring2017convolutional}. Notably, these advancements have contributed to the remarkable progress witnessed in large language models such as BERT~\cite{2018bert} and GPT~\cite{GPT2}. Motivated by these developments, attention-based spiking neural networks \cite{qin2023attention, TASNN} have recently been proposed to enhance the sequential modeling capabilities of SNNs. However, these models encounter challenges related to high computational complexity, limitations in hardware implementation, and the inability to operate in real-time scenarios. 

In this work, we present a novel spiking neuron model, dubbed \textbf{Contextual Embedding Leaky Integrate-and-Fire (CE-LIF)}. Our proposed model incorporates the concepts of adaptive firing threshold and position encoding to enhance the sequential modeling capability of SNNs. Specifically, we introduce a contextual embedding component to induce the firing threshold adjustments, which is produced by multiplying the membrane potential with a learnable temporal encoding (TE). Notably, the TE allows for the explicit representation of a neuron's preferred temporal pattern, thereby alleviating the challenge of preserving the temporal structure of the input signals within the adaptive firing threshold. To validate the effectiveness of the CE-LIF neuron, we conduct both theoretical analysis and extensive experimental studies, highlighting its superior sequential modeling capabilities. The key contributions of our work can be summarized as follows:
\begin{itemize}
\item We propose a novel spiking neuron model called CE-LIF that can significantly enhance the sequential modeling capability of SNNs.
\item We provide a theoretical analysis of the proposed CE-LIF neuron model. Our analysis offers valuable insights into how error gradients can be effectively propagated to distant inputs, thereby establishing long-term temporal dependencies.
\item We thoroughly validate the effectiveness of CE-LIF 
on a wide range of sequential modeling tasks. Notably, we demonstrate that the feedforward SNNs comprising CE-LIF neurons can surpass both state-of-the-art (SOTA) recurrent SNNs and LSTM models in terms of classification accuracy, network convergence speed, and memory capacity.
\end{itemize}

\section{Methods}

\begin{figure*}[htb]
\centering
\includegraphics[scale=0.49,trim= 160 10 0 20, clip]
{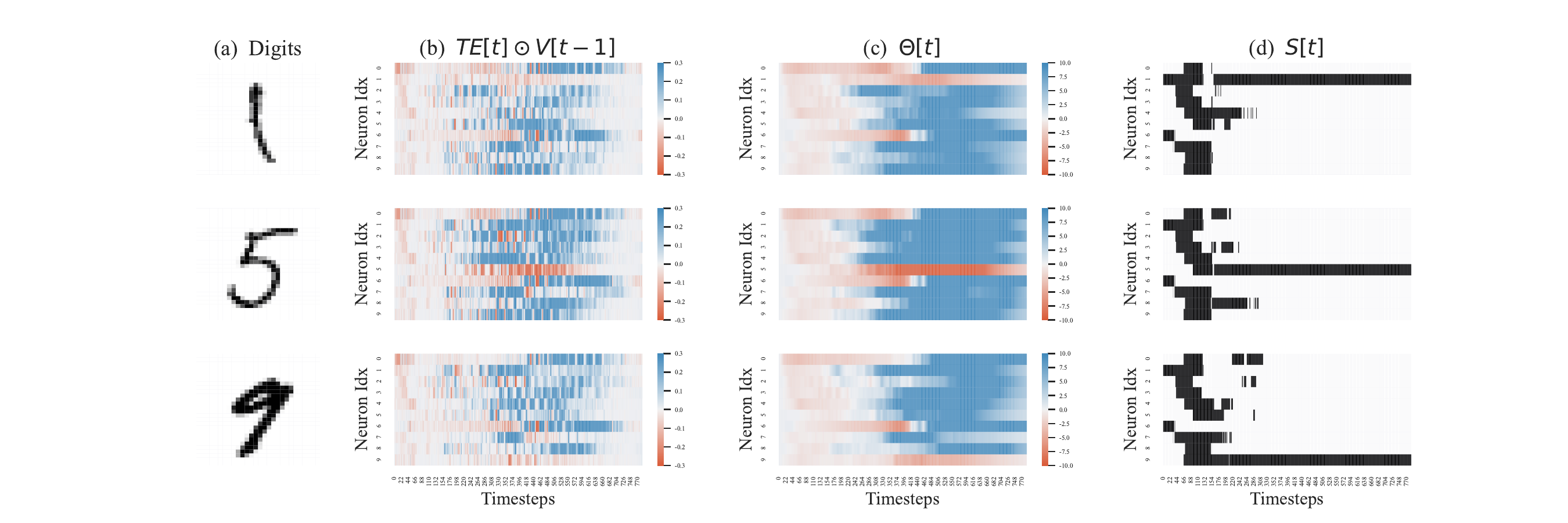}
\caption{Illustration of the evolution of neuronal variables during the Seq-MNIST classification task. \textbf{(a)} The input image is processed pixel-by-pixel through a raster scan. The recorded \textbf{(b)} contextual embeddings, \textbf{(c)} adaptive firing thresholds, and \textbf{(d)} the output spike trains of all output neurons. In (d), the black color bars represent the generated spikes.}
\label{fig: out_ana}
\end{figure*}

\subsection{Leaky Integrate-and-Fire (LIF) Neuron Model}
Inspired by the rich neuronal dynamics and firing behavior of biological neurons, a number of spiking neuron models have been proposed in the literature, offering varying levels of biological details \cite{neurons}. For large-scale SNN simulation, the leaky integrate-and-fire (LIF) model has commonly been adopted. This simplified neuron model can describe the essential membrane integration and spike generation processes of biological neurons while remaining computationally efficient. The neuronal dynamic of LIF neurons could be described as follows:
\begin{footnotesize}
\begin{equation}
\begin{aligned}
&\tau_m \frac{d v(t)}{d t}=-(v(t)-v_{reset})+I(t),\\
&\text{if} \quad v(t)\geq \theta, \quad v(t)\rightarrow v_{reset}.
\end{aligned}
\end{equation}
\end{footnotesize}
where $v_{reset}$ denotes the resting potential, $\tau_m$ is the membrane decaying time constant, and $I$ corresponds to the input current transduced from the input spikes $S(t)$, that is, $I(t) = \mathcal{W} S(t)$, where $\mathcal{W}$ is the synaptic weight. $\theta$ signifies the firing threshold. When the membrane potential reaches $\Theta$, it triggers the generation of an output spike. Following the spike generation, membrane potential will be reset to $v_{reset}$. In practice, this continuous-time formulation is usually discretized using the Euler method:
\begin{footnotesize}
\begin{equation}
\label{eq:lif}
\left\{\begin{array}{lr}
V[t]=\alpha\cdot V[t-1] \cdot (1-S[t-1])+I[t],& \\
S[t]=H\left(V[t]-\theta\right), &
\end{array}\right.
\end{equation}
\end{footnotesize}
where $\alpha=e^{-\tau_m}$ is the membrane potential decaying rate, $H(\cdot)$ is the Heaviside step function. SNNs comprising LIF neurons have demonstrated exceptional performance in pattern recognition tasks with limited temporal context. However, their suitability for tasks necessitating extended temporal dependencies is currently constrained due to two contributing factors. Firstly, the internal memory of LIF neurons, represented by the membrane potential, diminishes rapidly over time and is reset upon neuron firing. It, therefore, presents a significant challenge in retaining long-term memory. Secondly, the backpropagation-through-time (BPTT) algorithm, which is commonly employed in SNN training, poses an additional obstacle to learning long-term dependency. This is attributed to the recursive nature of gradient updates, which can lead to problems such as gradient vanishing and exploding. 

\subsection{Adaptive LIF Neuron Model}
Research on biological neurons has revealed that the neuronal firing threshold undergoes adaptations in response to neuronal firing activities. Specifically, an increase in the frequency of neuronal firing results in an elevation of the firing threshold \cite{sunkin2012allen}. This phenomenon has been identified as playing a crucial role in enhancing memory retention. Inspired by these findings, the adaptive threshold LIF (ALIF) neuron model has been introduced with the aim of enhancing the memory capacity of spiking neurons \cite{LSNN}. The formulation of the ALIF model can be summarized as follows:

\begin{footnotesize}
\begin{equation}
\left\{\begin{array}{lr}
\Theta[t]=\beta \cdot (\Theta[t-1]-\Theta_0) + \gamma \cdot S[t-1]+\Theta_0,&  \\
V[t]=\alpha \cdot V[t-1] \cdot (1-S[t-1])+I[t],& \\
S[t]=H\left(V[t]-\Theta[t]\right), &
\end{array}\right.
\end{equation}
\end{footnotesize}
where $\Theta_0$ represents the initial firing threshold. Each postsynaptic spike contributes to a transient increase in the firing threshold $\Theta$ with a gain of $\gamma$, whose contribution decays at a slow rate of $\beta$ (e.g., 0.99). This firing threshold adaptation mechanism encompasses two significant advantages. Firstly, the elevated firing threshold can effectively suppress excessive firing, thereby improving the efficiency of neural signal processing. Secondly, benefiting from its slow decay rate and non-resetting characteristics, the adaptive threshold can serve as a long-term memory. However, within such memory storage, the input information is compressed into low-resolution binary spikes with their temporal structures being inadequately represented, thereby presenting substantial obstacles when performing temporal credit assignments (see Supplementary Materials for more detailed analysis).

\subsection{Contextual Embedding LIF Neuron}
To address the above-identified deficiencies in ALIF neurons, we redesign the adaptive firing threshold by replacing the postsynaptic-spike-induced firing threshold increment with a contextual-embedding-induced one, we refer to the resulted neuron model as \textbf{Contextual Embedding LIF (CE-LIF)}. The refined adaptive firing threshold function is expressed as:
\begin{footnotesize}
\begin{equation}
\Theta[t]=\beta \cdot (\Theta[t-1]-\Theta_0)+TE[t] \odot V[t-1]+\Theta_0,
\label{CE-LIF}
\end{equation}
\end{footnotesize}
where $\odot$ denotes the Hadamard product. 
The proposed contextual embedding comprises two interacting components, namely the membrane potential $V$ and the temporal encoding vector $TE$. By replacing $S[t-1]$ with $V[t-1]$ in the firing threshold update function, the issue of low-resolution spike-based information representation is effectively addressed. This refinement not only aligns more closely with existing biological findings \cite{fontaine2014spike,zhang2003other}, but also facilitates the effective delivery of gradients to early timesteps (see analysis in Ablations and Supplementary Materials). Additionally, the temporal encoding vector $TE \in \mathbb{R}^{T}$ hard codes the preferred temporal pattern of a neuron, ensuring every neuron within the same layer possesses its unique TE vector. Therefore, the proposed contextual embedding effectively circumvents the challenge of preserving the temporal structure of input signals within the firing threshold. 

Notably, the inner product between $V$ and $TE$ acts as an indicator of the presence of the preferred temporal pattern. This information is subsequently integrated into the firing threshold to facilitate context-dependent processing. In order to elucidate the dynamic interaction between these two components in enabling context-dependent processing, we provide a visualization in Figure \ref{fig: out_ana} to illustrate the evolution of $TE \odot V$, $\Theta$, and $S$ throughout the Sequential MNIST (Seq-MNIST) classification task.
In this task, input images from the MNIST dataset are processed on a pixel-by-pixel basis via a raster scan \cite{smnist}. Notably, as depicted in Figure \ref{fig: out_ana}(b), the $TE \odot V$ of target neurons often display negative values, indicating the detection of their preferred temporal patterns. Consequently, this leads to a consistent reduction in the firing threshold $\Theta$ in Figure \ref{fig: out_ana}(c), ultimately resulting in the observed bursting behaviors exhibited by the target neurons, as illustrated in Figure \ref{fig: out_ana}(d). In contrast, non-target neurons more frequently exhibit positive values in their $TE \odot V$ representation, leading to elevated firing thresholds that inhibit them from firing. 

\subsection{Impact of CE on Gradient Propagation}

Here, we provide a theoretical analysis to explain how gradients can be effectively delivered to earlier timesteps in CE-LIF neurons. Specifically, we adopt the BPTT algorithm along with surrogate gradient \cite{wu2018spatio,8891809} for parameter update, which can be summarised as follows:
\begin{scriptsize}
\begin{equation}
\begin{aligned}
\label{eq:gradw}
 &\Delta \mathcal{W}^{l} \propto \frac{\partial \mathcal{L}}{\partial \mathcal{W}^{l}}= \sum_{t=1}^{T}{\frac{\partial \mathcal{L} }{\partial V^{l}[t]}S^{l-1}[t]},
 &\Delta b^{l} \propto \frac{\partial \mathcal{L} }{\partial b^{l}}= \sum_{t=1}^{T}{\frac{\partial \mathcal{L} }{\partial V^{l}[t]}},
\end{aligned}
\end{equation}
\end{scriptsize}
where $\mathcal{L}$ is the loss function. $\mathcal{W}^l$ and $b^l$ refer to weight and bias terms of layer $l$, respectively. The partial derivative of $\frac{\partial \mathcal{L} }{\partial V^l[t]}$ can be further decomposed and calculated as follows: 
\begin{scriptsize}
\begin{equation}
\label{eq:grad}
\begin{aligned}
\frac{\partial \mathcal{L} }{\partial V^l[t]}& =\frac{\partial \mathcal{L} }{\partial S^l[t]} \frac{\partial S^l[t]}{\partial V^l[t]}+\frac{\partial \mathcal{L} }{\partial V^l[{\scriptstyle t+1}]} \frac{\partial V^l[t+1]}{\partial V^l[t]} +\frac{\partial \mathcal{L} }{\partial \Theta^l[{\scriptstyle t+1}]} \frac{\partial \Theta^l[t+1]}{\partial V^l[t]} \\   
& = \frac{\partial \mathcal{L} }{\partial S^l[t]}g^{\prime}[t] +\frac{\partial \mathcal{L} }{\partial V^l[{\scriptstyle t+1}]} \alpha(1-S^l[t]) +\frac{\partial \mathcal{L} }{\partial \Theta^l[{\scriptstyle t+1}]} TE[t+1],\\
\frac{\partial \mathcal{L} }{\partial S^l[t]}&=\frac{\partial \mathcal{L} }{\partial V^l[t+1]} \frac{\partial V^l[t+1]}{\partial S^l[t]}+\frac{\partial \mathcal{L} }{\partial V^{l+1}[t]} \frac{\partial V^{l+1}[t]}{\partial S^l[t]} \\
&=-\frac{\partial \mathcal{L} }{\partial V^l[t+1]}\alpha V^l[t]+\frac{\partial \mathcal{L} }{\partial V^{l+1}[t]} W^{l},   \\
 \frac{\partial \mathcal{L} }{\partial \Theta^l[t]}&=\frac{\partial \mathcal{L} }{\partial S^l[t]} \frac{\partial S^l[t]}{\partial \Theta^l[t]}+\frac{\partial \mathcal{L} }{\partial \Theta^l[t+1]} \frac{\partial \Theta^l[t+1]}{\partial \Theta^l[t]} \\
 &=-\frac{\partial \mathcal{L} }{\partial S^l[t]} g^{\prime}[t]+\frac{\partial \mathcal{L} }{\partial \Theta^l[t+1]} \beta,
\end{aligned}
\end{equation}
\end{scriptsize}
where $g^{\prime}[t]=H^{\prime}(V^l[t]-\Theta^l[t])$ refers to the surrogate gradient function, which is applied during error backpropagation to address the discontinuity of spike generation function given in Eq. (\ref{eq:lif}). In particular, we adopt a boxcar function \cite{wu2018spatio} that is defined as follows:
\begin{footnotesize}
\begin{equation}
\label{eq:sg}
   g^{\prime}[t] = \left\{\begin{array}{lr} 1, \quad \text{if} \quad |V^l[t]-\Theta^l[t]|<\Gamma,\\ 0,  \quad \text{otherwise,} \end{array}\right.
\end{equation}
\end{footnotesize}
where $\Gamma$ is a hyperparameter that determines the permissible range for gradient propagation.

\begin{figure}[t]
\centering
\includegraphics[scale=0.34,trim= 0 0 0 0, clip]{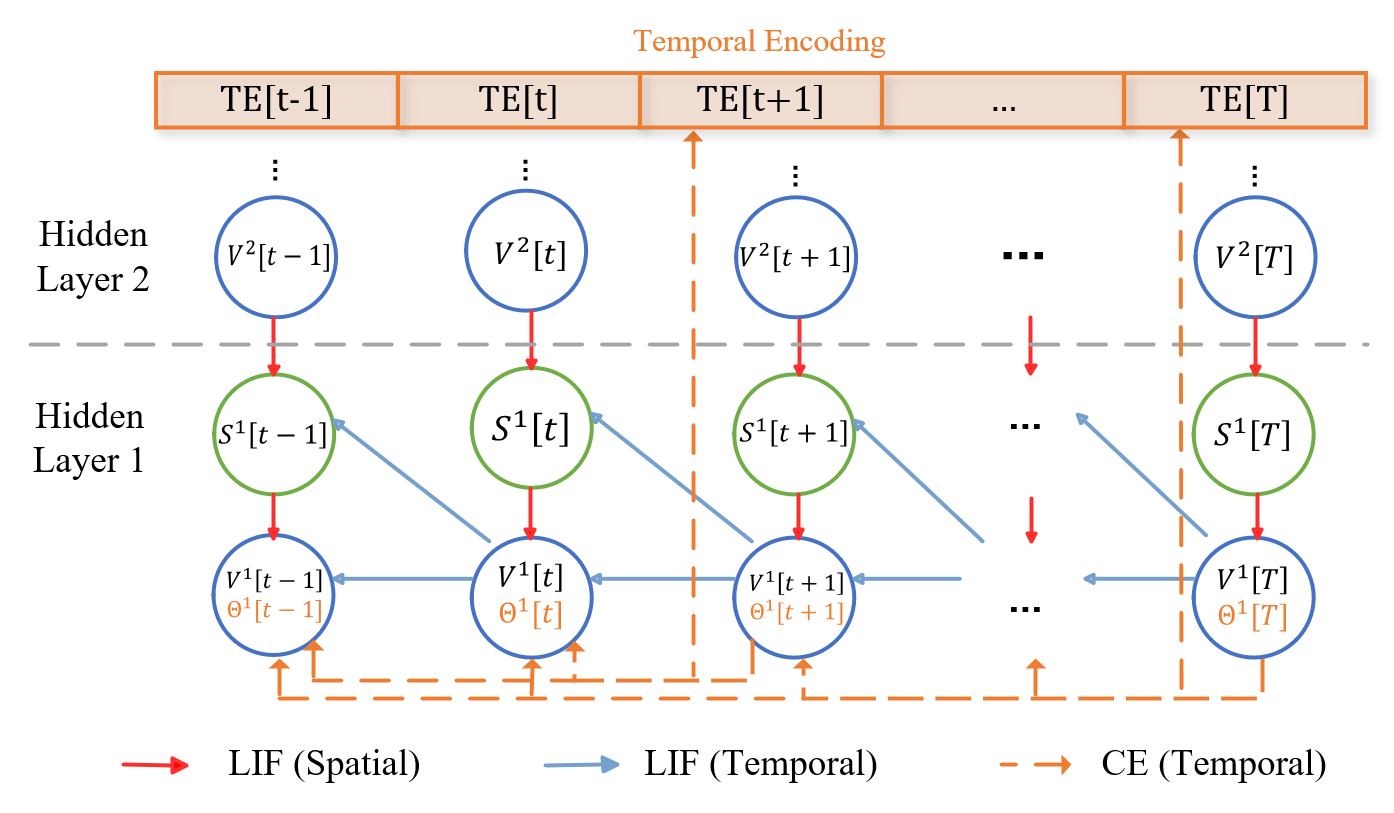}
\caption{ Illustration of the error backpropagation in an SNN comprising CE-LIF neurons. The backward paths of CE (orange dotted lines) establish a highway that can backpropagate the error information directly to earlier timesteps. Better to view this figure in color.}
\label{fig: ff_fb}
\end{figure}

Subsequently, by transforming Eq. (\ref{eq:grad}) into a cumulative form using recursive accumulation, we have:
\begin{footnotesize}
\begin{equation}
\begin{aligned}
\label{eq:gradth2}
 \frac{\partial \mathcal{L} }{\partial \Theta^l[t]}&=\sum_{i=t}^{T}{ {-\frac{\partial \mathcal{L} }{\partial S^l[i]}g^{\prime}[t]\beta^{i-t}}}=\sum_{i=t}^{T}{ {-\frac{\partial \mathcal{L} }{\partial S^l[i]}\frac{\partial S^l[i]}{\partial V^l[i]}\beta^{i-t}}} \\
 &= \sum_{i=t}^{T}{ {-\frac{\partial \mathcal{L} }{\partial V^l[i]}\beta^{i-t}}},
\end{aligned}
\end{equation}
\end{footnotesize}

Finally, by substituting Eq. (\ref{eq:gradth2}) into Eq. (\ref{eq:grad}), we can rewrite $\frac{\partial \mathcal{L} }{\partial V^l[t]}$ as follows:

\begin{footnotesize}
\begin{equation}
\begin{aligned}
\label{eq:gradv2}
\frac{\partial \mathcal{L} }{\partial V^l[t]} =&\underbrace{ \frac{\partial \mathcal{L} }{\partial S^l[t]} g^{\prime}[t]}_{\text{LIF (Spatial)}}+\underbrace{\frac{\partial \mathcal{L} }{\partial V^l[t+1]} \alpha (1-S^l[t])}
_{\text{LIF (Temporal)}}\\
&- \underbrace{\sum_{i=t+1}^{T}{{\frac{\partial \mathcal{L} }{\partial V^l[i]}}} \beta^{i-t} TE[t+1]}_{\text{CE}},
\end{aligned}
\end{equation}
\end{footnotesize}

\begin{table*}[!t]
\caption{Comparison of model performance on pixel-level sequential image classification tasks. For both Seq-MNIST and PS-MNIST datasets, the sequence lengths are set to 784.}
\label{sample-table}
\begin{center}
\resizebox{0.85\textwidth}{!}{
\begin{tabular}{cccccc}
\hline
\multicolumn{1}{c}{\bf Model} &\multicolumn{1}{c}{\bf SNN} &\multicolumn{1}{c}{\bf Architecture} &\multicolumn{1}{c}{\bf Parameters} &\multicolumn{1}{c}{\bf Seq-MNIST} &\multicolumn{1}{c}{\bf PS-MNIST}  \\
\hline
 LSTM \cite{URNN} & No & Recurrent & 66.5k & 98.20\% & 88.00\% \\
 IndRNN \cite{indrnn} & No & Recurrent & 99.2k & 99.50\% & 96.00\% \\
 TCN \cite{TCN} & No & Convolution & 70.0k & 99.00\% & 97.20\% \\
 \hline
 LSNN \cite{LSNN} & Yes & Recurrent & 68.2k  & 93.70\% & N.A. \\
 DEEP-R LSNN \cite{LSNN} & Yes & Recurrent & 66.4k  & 96.40\% & N.A. \\
 DEXAT \cite{DEXAT}& Yes & Recurrent & N.A. & 96.40\% & N.A. \\
 AHP \cite{AHP}& Yes & Recurrent & 68.4k & 96.00\% & N.A. \\
 LIF* & Yes & Feedforward & 85.1k & 63.45\% & - \\
 LTMD* \cite{LTMD}& Yes & Feedforward & 85.1k  & 68.56\% & - \\
 PLIF* \cite{PLIF}& Yes & Feedforward & 85.1k & 87.92\% & - \\
 GLIF* \cite{GLIF}& Yes & Feedforward & 85.4k & 95.27\% & - \\
  \textbf{CE-LIF (Ours)}& \textbf{Yes} & \textbf{Feedforward} &\textbf{83.5k} &\textbf{98.19\%} &\textbf{97.90\%} \\
\hline
 LIF* & Yes & Recurrent & 155.1k & 84.14\% &55.13\% \\ 
 LTMD* \cite{LTMD}& Yes & Recurrent & 155.1k &  84.62\% & 54.93\% \\
 ALIF \cite{ALIF}& Yes & Recurrent & 156.3k & 98.70\% & 94.30\% \\ 
PLIF* \cite{PLIF}& Yes & Recurrent & 155.1k & 91.79\% & - \\
 GLIF* \cite{GLIF}& Yes & Recurrent & 157.5k & 96.64\% & 90.47\% \\
 \textbf{CE-LIF (Ours)} & \textbf{Yes} & \textbf{Feedforward} &\textbf{153.9k} &\textbf{98.84\%} &\textbf{98.14\%} \\ 
 \hline
 \multicolumn{2}{l}{* \hspace{5.6mm} Our reproduced results based on  publicly available codebases} \\
 \multicolumn{1}{l}{- \hspace{6.2mm} These models are unable to converge} \\
 \multicolumn{1}{l}{N.A. \hspace{0.7mm} These results are not publicly available}
\label{tab:mnist}
\end{tabular}}
\vspace{-3mm}
\end{center}
\end{table*}

where the first two terms have an identical format as in the LIF model. The first term, as illustrated in Figure \ref{fig: ff_fb}, represents the spatial credit assignment (SCA) between consecutive layers. However, the SCA may encounter the dead gradient problem when the surrogate gradient value $g^{\prime}[t]$ equals zero. Furthermore, the second term corresponds to the temporal credit assignment (TCA) that decays exponentially at a rate of $\alpha$ as well as resets after each spike generation. Consequently, the gradient information cannot be effectively propagated over a long period of time. Notably, our proposed CE can effectively address this notorious problem by establishing a gradient highway between all dependent timesteps. By incorporating appropriately learned values of ${TE}$ alongside a slow decay rate of $\beta$, the CE-LIF model can effectively propagate gradients to much more distant timesteps compared with the ALIF model, as will be demonstrated in the Experiments section. 

\section{Experiments}
In this section, we provide a comprehensive analysis of the proposed CE-LIF model in terms of the sequential modeling capability, network convergence speed, and efficacy in performing long-term TCA. Moreover, ablation studies among different implementations of contextual embedding are also provided to demonstrate the superiority of the proposed design. The detailed experimental setups are provided in the Supplementary Materials. 

\subsection{Pixel-level Sequential Image Classification}

\begin{figure*}[!htb]
\centering
\subfigure[]{
	\includegraphics[trim=0 0 0 20,clip,scale=0.28]{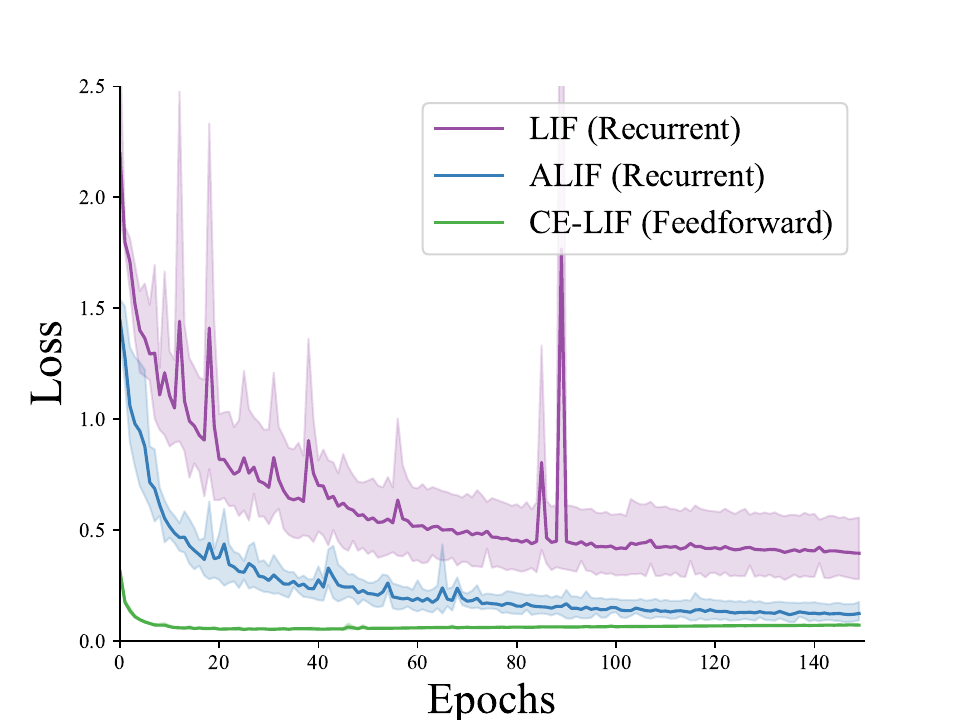}}\hspace{0em} 
\subfigure[]{
	\includegraphics[trim=10 0 78 30,clip,scale=0.28]{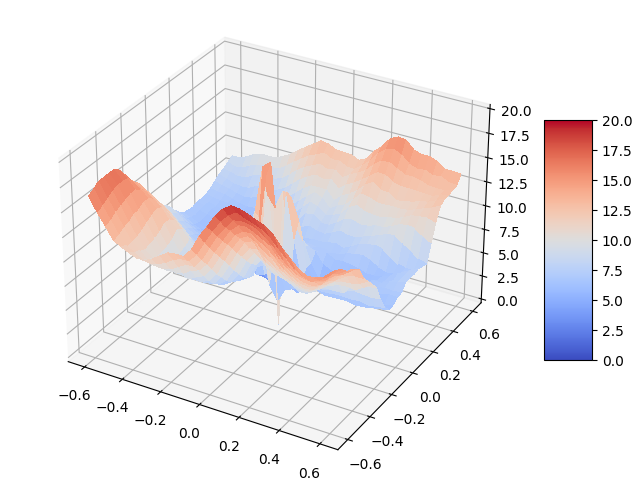}}\hspace{0em}
\subfigure[]{
	\includegraphics[trim=10 0 78 30,clip,scale=0.28]{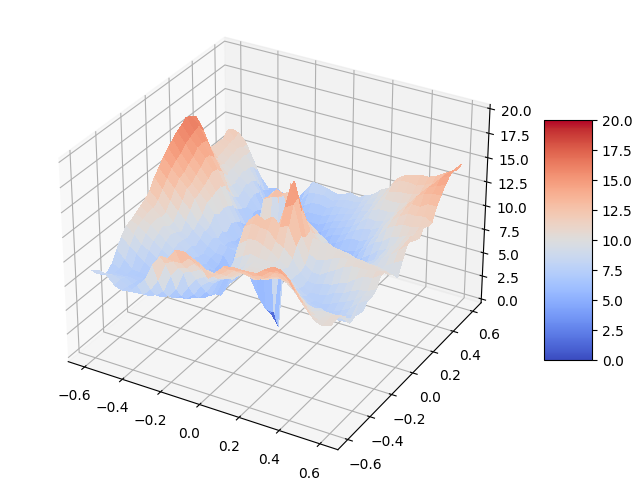}}\hspace{0em}
\subfigure[]{
	\includegraphics[trim=10 0 6 30,clip,scale=0.28]{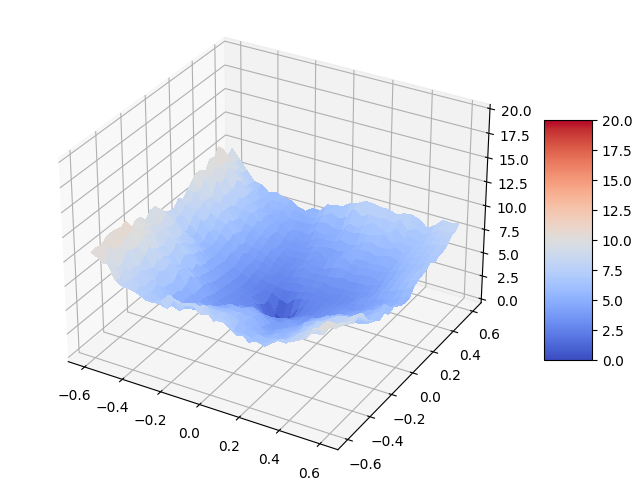}}\hspace{0em}
\caption{Comparison of \textbf{(a)} learning curves and \textbf{(b-d)} loss landscapes among LIF, ALIF, and CE-LIF models. Note that the CE-LIF model adopts a feedforward structure, while LIF and ALIF models use a recurrent one. These structures are chosen to maximize the respective model's performance. The results in (a) are obtained across three runs with different random seeds. }
\label{fig: loss}
\vspace{-3mm}
\end{figure*}

\begin{figure*}[t]
\centering
\subfigure[]{
	\includegraphics[scale=0.28, trim=15 20 30 10, clip]{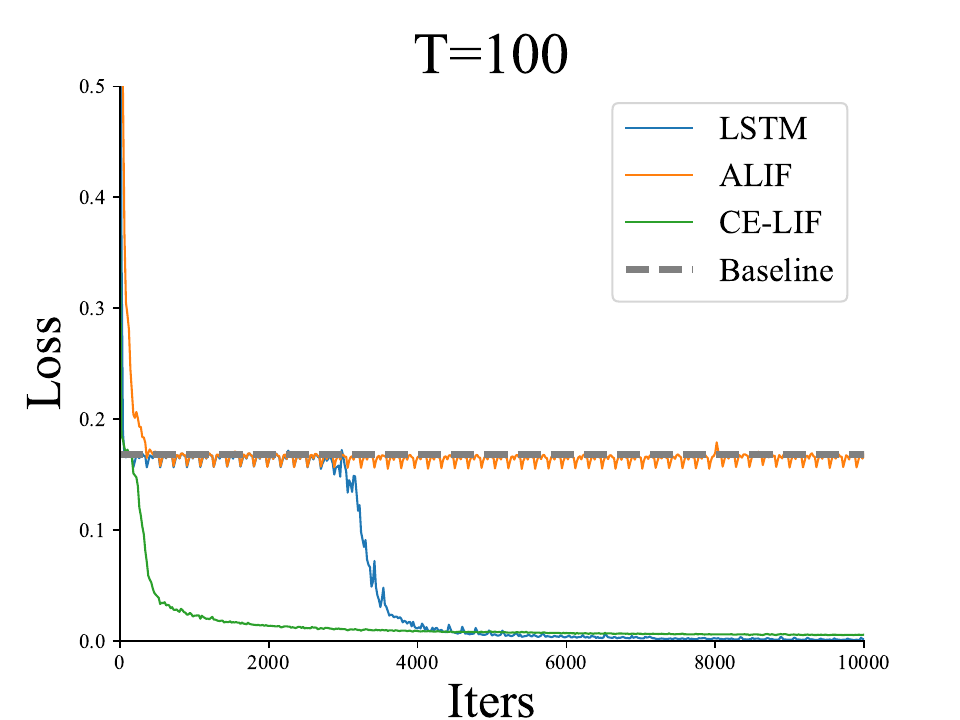}}\hspace{0em}
\subfigure[]{
	\includegraphics[scale=0.28, trim=35 20 30 10, clip]{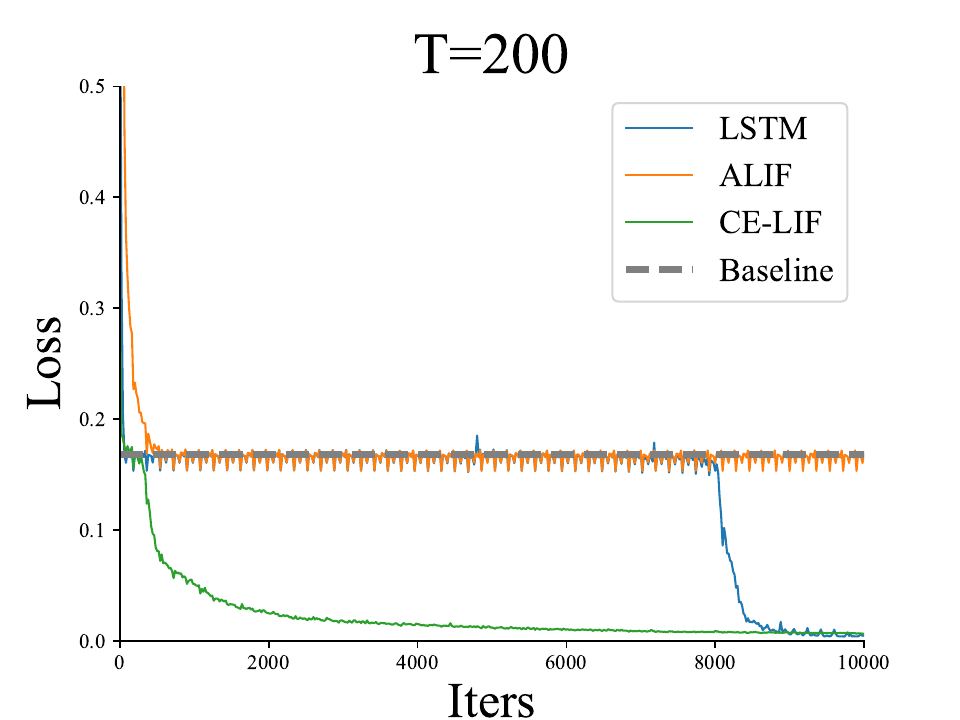}}\hspace{0em}
\subfigure[]{
	\includegraphics[scale=0.28, trim=35 20 30 10, clip]{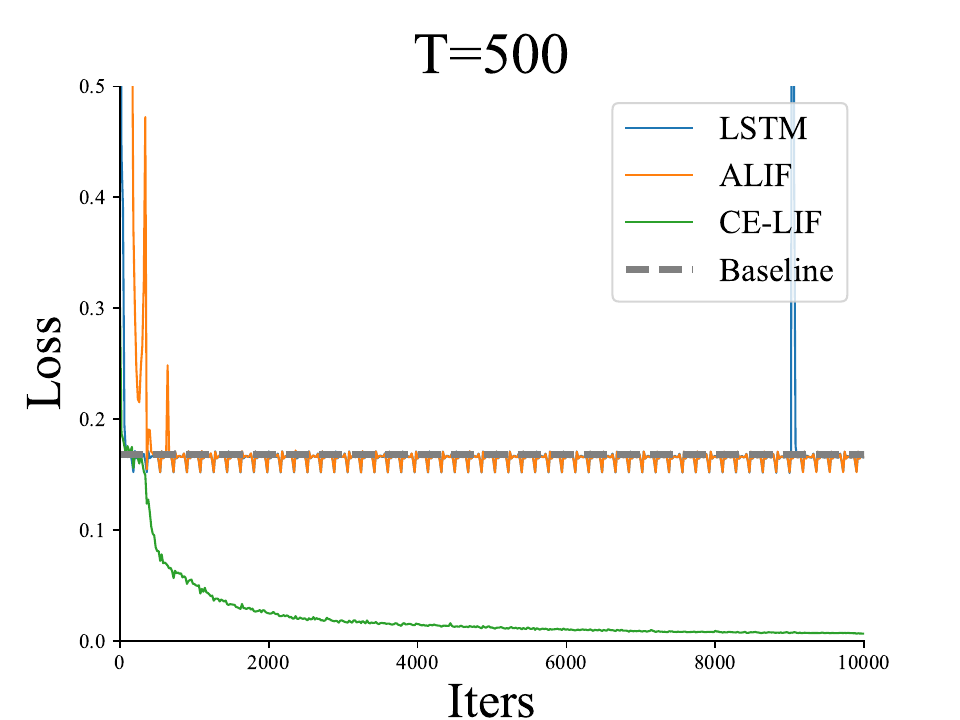}}\hspace{0em}
\subfigure[]{
	\includegraphics[scale=0.28, trim=35 20 30 10, clip]{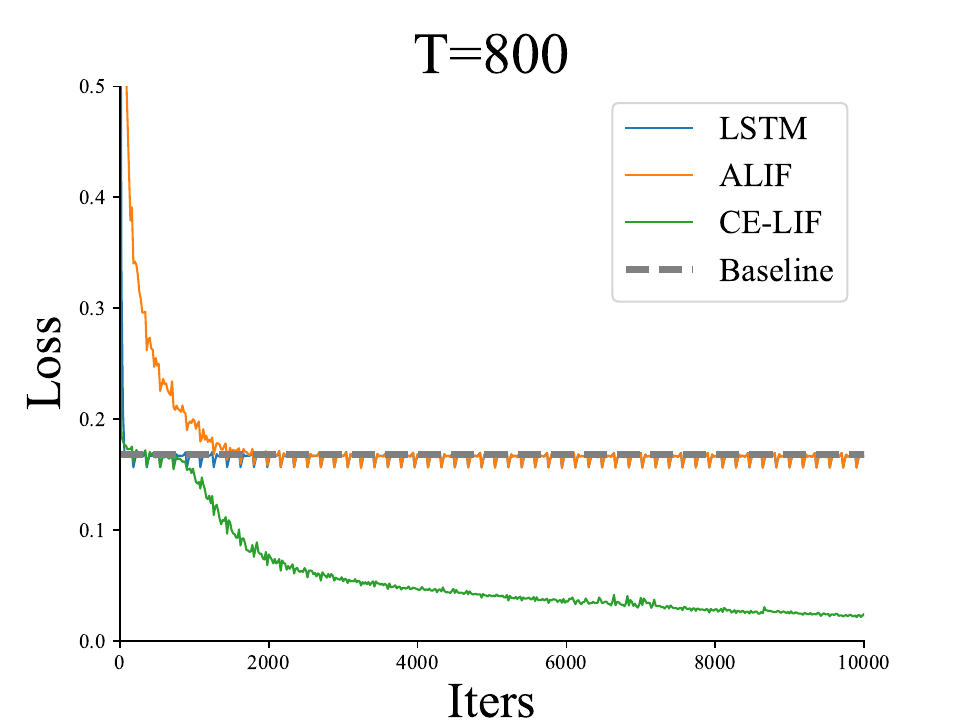}}\hspace{0em}\\

\subfigure[]{
	\includegraphics[scale=0.28, trim=15 0 35 10, clip]{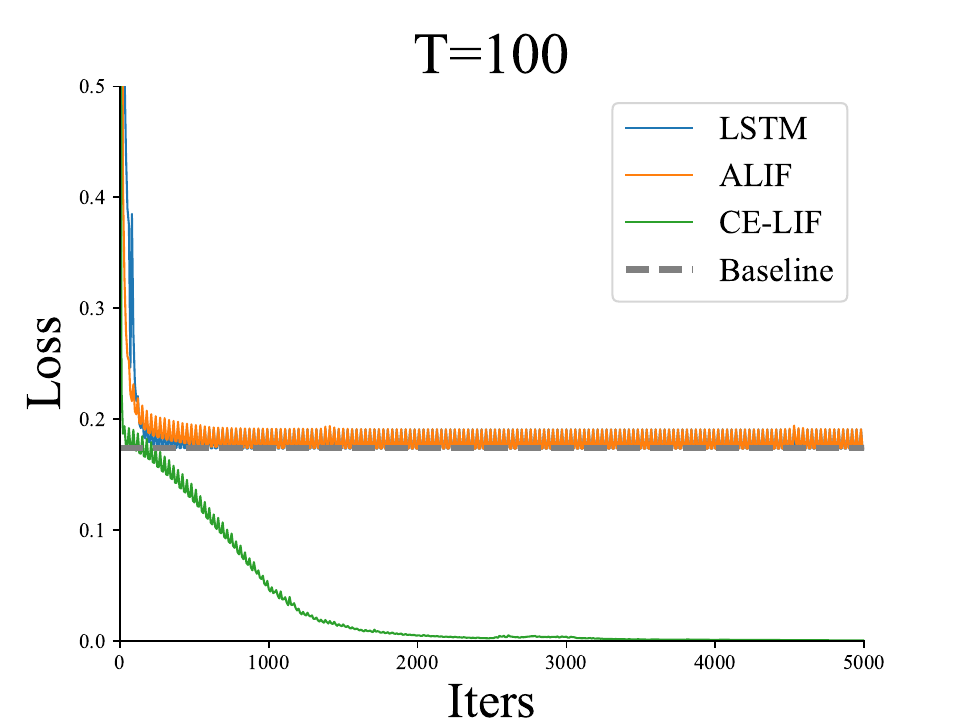}}\hspace{0em}
\subfigure[]{
	\includegraphics[scale=0.28, trim=33 0 35 10, clip]{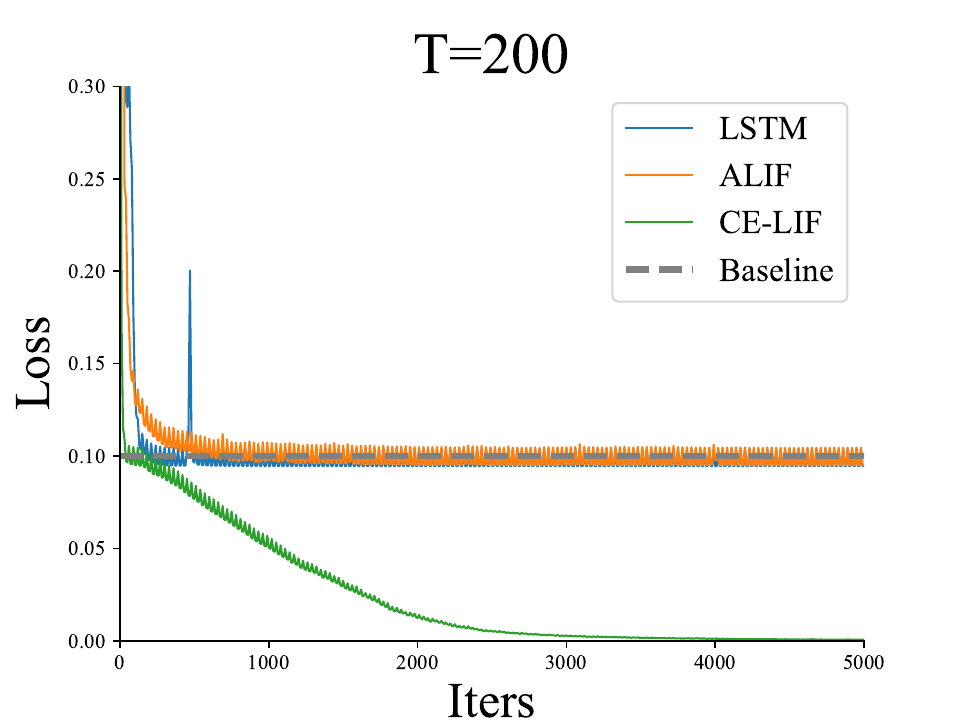}}\hspace{0em}
\subfigure[]{
	\includegraphics[scale=0.28, trim=33 0 35 10, clip]{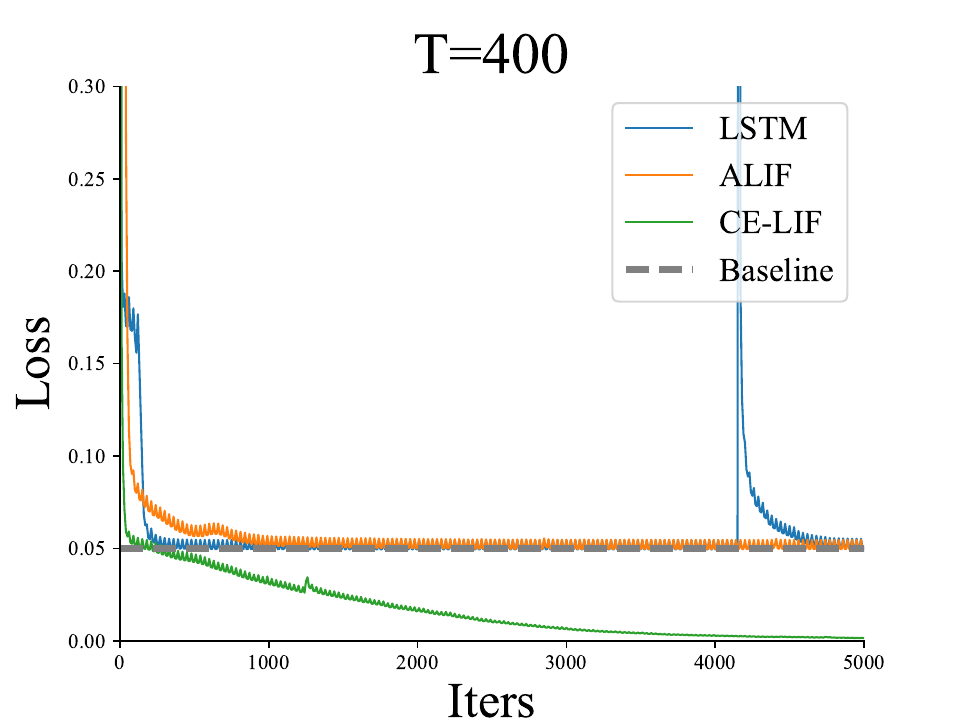}}\hspace{0em}
\subfigure[]{
	\includegraphics[scale=0.28, trim=33 0 35 10, clip]{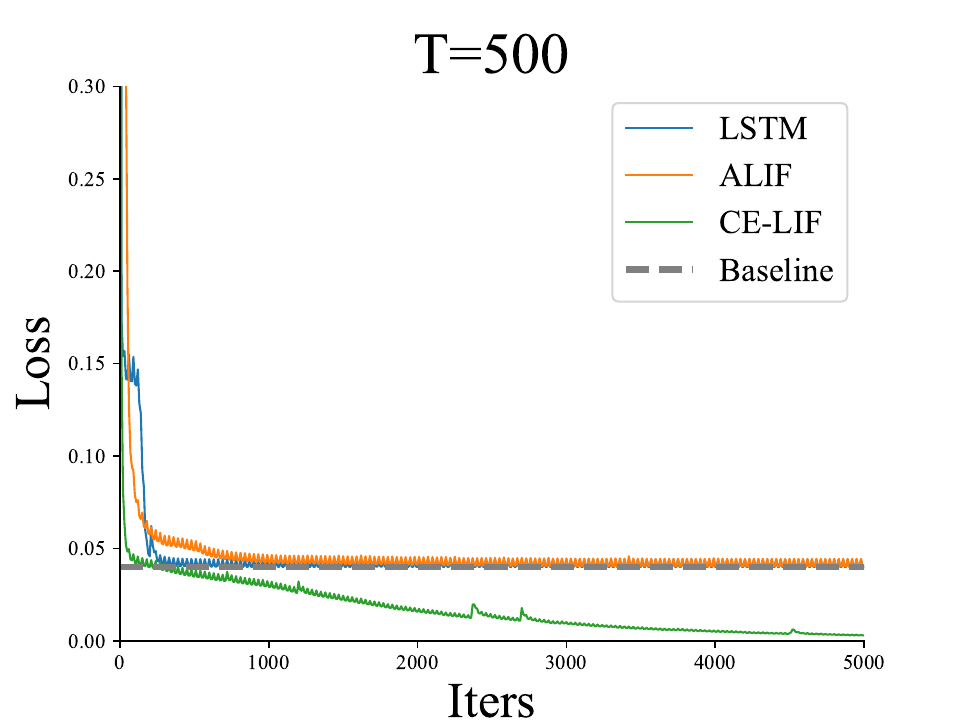}}\hspace{0em}
 \vspace{-3mm}
\caption{Learning curves for \textbf{(a-d)} adding problem and \textbf{(e-h)} copy memory problem. }
\label{fig: add_copy}
\end{figure*}

We first evaluate the performance of our proposed CE-LIF model on two widely adopted sequential classification tasks, namely Seq-MNIST and Permuted Sequential MNIST (PS-MNIST) \cite{smnist}. In these tasks, the input images are processed by a model pixel-by-pixel through a raster scan. As the results summarized in Table \ref{tab:mnist}, the CE-LIF model outperforms all other SOTA SNN implementations with the same or a lesser number of parameters. These results are also competitive with other ANN models, such as LSTM~\cite{URNN}, IndRNN~\cite{indrnn}, and TCN~\cite{TCN}. Notably, the results of our CE-LIF model are achieved with a feedforward network structure, while almost all other works employ a recurrent network structure to achieve effective sequential modeling. Considering that our feedforward SNNs have already achieved saturation in terms of classification accuracy on these datasets, introducing a recurrent network structure would lead to an unnecessary parameter increase. As a result, we restrict our analysis of CE-LIF to a feedforward SNN in the following sections. For a more comprehensive analysis of the spiking RNN utilizing CE-LIF neurons, we refer readers to the Supplementary Materials.

Moreover, the CE-LIF model exhibits significantly faster and more stable training convergence compared with LIF and ALIF models. As demonstrated in Figure \ref{fig: loss}, the CE-LIF model achieves rapid convergence on the Seq-MNIST task within a mere ten training epochs, whereas the LIF and ALIF models require more than 100 epochs to converge. Furthermore, the CE-LIF model exhibits a much smoother loss landscape, which facilitates training convergence and offers great potential for improved generalization. These discrepancies do not primarily stem from the ease of convergence in a feedforward structure. More crucially, they can be attributed to the more effective TCA established by the proposed contextual embedding. Therefore, these findings highlight the superiority of the proposed CE-LIF model in sequential modeling.

\subsection{Memory Capacity Analysis}
\label{seq_model_tasks}
We further analyze the memory capacity of the proposed CE-LIF model on the adding problem and copy memory problem that necessitates long-term memory \cite{LSTM97}. Specifically, the comparative study is performed among the CE-LIF, LSTM, and ALIF models. 

\textbf{Adding problem} is a regression task in which two input sequences of length $T$ are provided, one representing a value sequence and the other an indicator sequence \cite{URNN}. The value sequence comprises randomly sampled entries from the range of $(0,1)$, while the indicator sequence assigns a value of $1$ to two randomly selected entries and $0$ to the remaining entries. The indicator sequence works as a pointer to indicate which two entries in the value sequence should be added. In this task, input sequences are processed by models one entry after the other, and the model will output the sum after receiving all entries. 
The mean squared error (MSE) is used as the loss function $\mathcal{L}$. It is worth noting that the baseline MSE for this task is 0.167, obtained by assuming both entries take an average value of 0.5. The learning curves for the sequence lengths of $T=100$, $200$, $500$, and $800$ is presented in Figure \ref{fig: add_copy}(a-d). We notice that feedforward SNNs comprising CE-LIF neurons can successfully converge across all sequence lengths. In contrast, the selected baseline models exhibit inferior performance, with LSTM performing better than recurrently connected ALIF. Despite the slower convergence speed, the LSTM model reaches an MSE comparable to that of the CE-LIF model when $T=100$ and $200$. However, both models fail to converge for longer sequence lengths. Notably, the CE-LIF model exhibits a slight increase in MSE with increasing sequence length, attributed to the accumulated quantization errors from spike-based representation.

\begin{figure}[h]
\centering
\includegraphics[scale=0.39,trim= 30 0 10 30, clip]{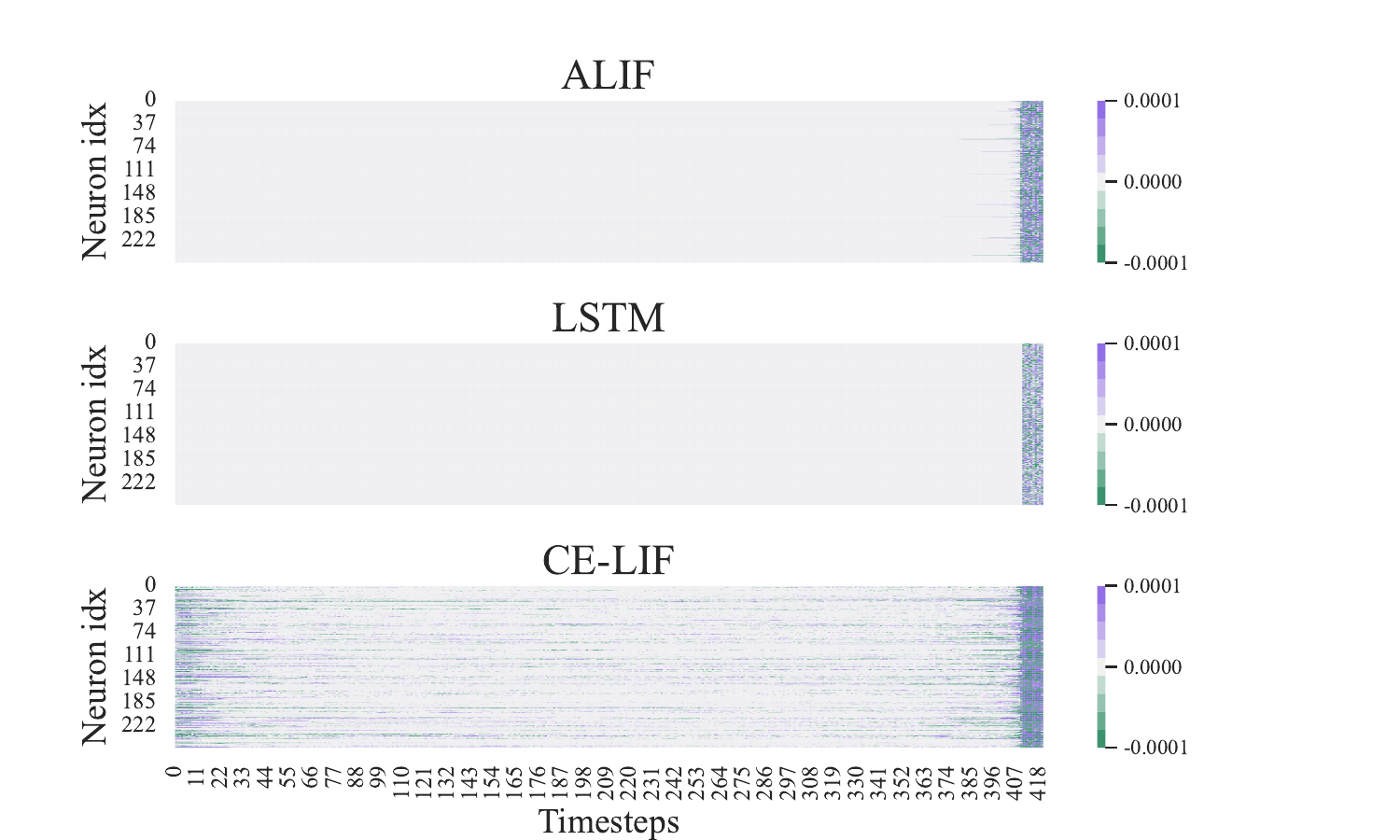}
\caption{Illustration of the $Grad[t]$ calculated for the ALIF, LSTM, and CE-LIF models in performing long-term TCA.} 
\label{fig: grad}
\end{figure}

\textbf{Copy memory problem} is a classification task with 10 output classes ranging from $0$ to $9$. The total input sequence length is $T+20$, where the key input sequence $x_{0:10}$ is fed into the model during the first $10$ timesteps, with each entry uniformly sampled from integer $1$ to $8$. For the following $T$ timesteps, the inputs are set to $0$. During the last $10$ timesteps, the model consistently receives an input of $9$, signifying the decision-making phase, and the model is expected to output $x_{0:10}$ sequentially during this period. To successfully accomplish this task, the model is anticipated to possess long-term memory that can hold the input $x_{0:10}$ for a prolonged period of $T$. To match the spike-based representation in SNNs, we employ one-hot encoding to encode the input values before feeding them into SNNs, with cross-entropy (CE) loss being used as the loss function. Assuming $y_{0:T+10}=0$ and $y_{T+10:\text{end}}$ are sampled uniformly at random from $0$ to $9$, the baseline loss can be calculated as $\frac{10\ln({8})}{T+20}$ \cite{URNN}. Figure \ref{fig: add_copy} (e-h) present the learning curves for $T=100$, $200$, $400$, and $500$. Remarkably, the feedforward SNNs comprising the proposed CE-LIF neurons can converge within 5,000 iterations, even for the challenging case of $T=500$. In contrast, both the LSTM and ALIF models fail to surpass the baseline value even with the shortest time interval $T=100$, indicating they can not preserve information beyond the $T$ time interval.

\subsection{Efficacy of CE in Facilitating Long-term TCA}
To provide more insights into how long-term temporal relationships have been established in CE-LIF neurons, we provide a visualization in Figure \ref{fig: grad} to illustrate the error gradients calculated for CE-LIF, LSTM, and ALIF models during training. All models have been trained on the copy memory problem with $T=400$, and the gradients $Grad[t]=\partial \mathcal{L}/\partial S^l[t]$ of the last hidden layer are reported. In this task, the inputs that contain the most valuable information are located within the first $10$ timesteps, while the memory recall occurs during the last $10$ timesteps. Therefore, an ideal model for this task should be able to establish long-term dependencies between the output error and the distant inputs that are situated $410$ timesteps away. Noticeably, the CE-LIF model demonstrates an impressive ability to backpropagate gradients to these distant inputs effectively. In contrast, both the ALIF and LSTM models encounter challenges in propagating gradients across such a lengthy interval, although the ALIF model exhibits slightly better performance compared to the LSTM model. This empirical observation aligns with the theoretical analysis presented in the earlier section.

\subsection{Ablation Study of Contextual Embedding Design}

In the CE-LIF model, the contextual embedding is formulated as the product of $TE$ and $V$. Here, we conduct an ablation study to evaluate the effectiveness of this design. The results presented in Table \ref{tab:abl} demonstrate that the proposed design outperforms alternative candidates on both benchmarks. Notably, a significant improvement is observed when changing from $S[t-1]$ used in ALIF to $V[t-1]$. Furthermore, the accuracy is further improved when multiplying $V[t-1]$ with the temporal encoding $TE[t]$ that facilitates the temporal structure modeling. Hence, this study provides compelling evidence regarding the significant effectiveness of the proposed contextual embedding design. 
\begin{table}[h]
\caption{Comparison of various contextual embedding options in sequential image classification tasks. Note that a feedforward SNN is adopted in this experiment.}
\begin{center}
\resizebox{0.9\textwidth}{!}{
\begin{tabular}{ccccccc}
\hline
\multicolumn{1}{c}{\bf ID}  & 
\multicolumn{1}{c}{$\bf CE[t]=$}  & 
\multicolumn{1}{c}{\textbf{Seq-MNIST}}  & \multicolumn{1}{c}{\textbf{PS-MNIST}} \\
\hline
1 & $ S[t-1]$ & 68.66\% & 36.64\% \\ 

2 & $ V[t-1]$ & 87.55\% & 72.60\% \\ 

3 & $ TE[t]$ & 95.36\% & 90.19\% \\

4 & $ V[t-1]+TE[t]$ & 95.55\%  &  89.97\%\\

5 & $ S[t-1]\odot TE[t]$ & 98.61\% &  96.21\% \\

6 & $\bf V[t-1]\odot TE[t]$ & \textbf{98.84\%}  &  \textbf{98.14\%}  \\
\hline
\label{tab:abl}
\end{tabular}}
\vspace{-4mm}
\end{center}
\end{table}

\subsection{Visualization of Temporal Encoding}
\begin{figure}[t]
\centering

\subfigure[]{
	\includegraphics[scale=0.35,trim= 70 20 90 35, clip]{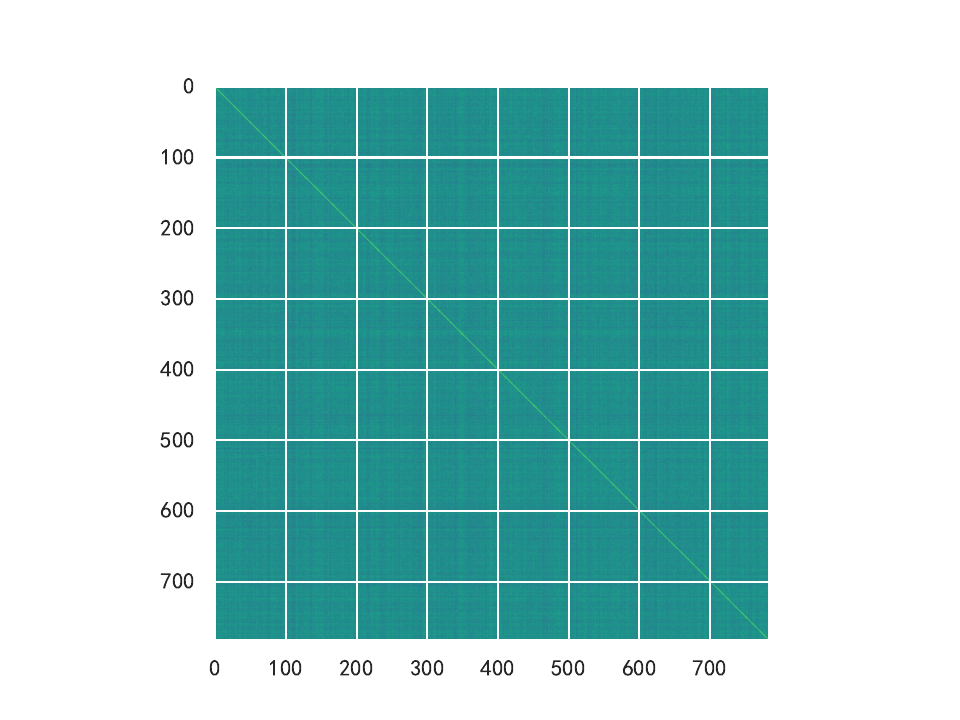}}\hspace{0em}
 \subfigure[]{
	\includegraphics[scale=0.35,trim= 70 20 40 35, clip]{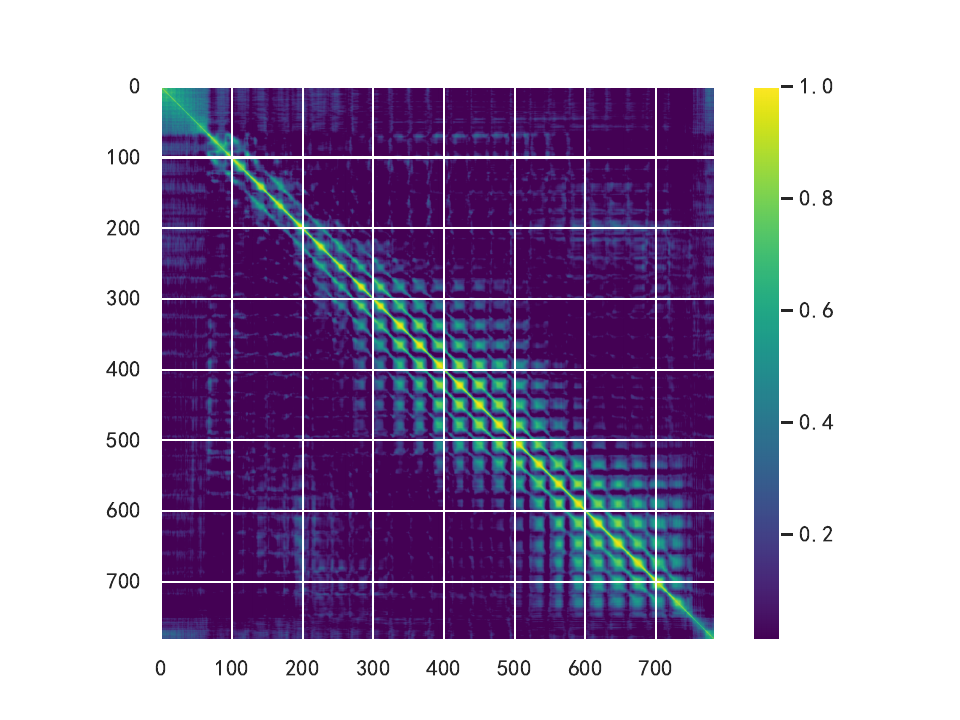}}\hspace{0em}
\caption{
The cosine similarity between different time steps of $TE$ \textbf{(a)} before training and \textbf{(b)} after training.}
\vspace{-3mm}
\label{fig: whatlearn}
\end{figure}

While previous sections have highlighted the superior sequential modeling capabilities of the CE-LIF model, it remains unclear what information is stored in TE that contributes to effective sequential modeling. Therefore, we investigate the learned TE on the Seq-MNIST task. As the heat map presented in Figure \ref{fig: whatlearn}, the color at each coordinate $(i,j)$ represents the cosine similarity between $TE[i]$ and $TE[j]$. Before training, there is no transparent correlation between different time steps. In contrast, the heat map exhibits periodic patterns after training, demonstrating higher similarities between neighboring time steps and time steps that are multiples of 28 apart. This pattern aligns well with the structure of the input image, suggesting that $TE$s contribute to the modeling of temporal structure. Furthermore, it is observed that later time steps exhibit broader connections with earlier ones, thus contributing to effective decision-making by incorporating a broader temporal context.

\section{Conclusion}
In this work, we introduced a novel spiking neuron model called CE-LIF, which can enhance the sequential modeling capabilities of SNNs. The theoretical analysis and extensive experimental studies demonstrated the efficacy of the proposed contextual embedding in facilitating long-term TCA, thereby leading to rapid network convergence and superior temporal classification accuracies. Furthermore, our experiments on the copy memory problem demonstrated the significantly increased memory capacity of CE-LIF compared to SOTA ALIF and LSTM models. It, therefore, opens up a myriad of opportunities for solving challenging temporal signal processing tasks with neuromorphic solutions.


\begin{thebibliography}{38}
\providecommand{\natexlab}[1]{#1}

\bibitem[{Arjovsky, Shah, and Bengio(2016)}]{URNN}
Arjovsky, M.; Shah, A.; and Bengio, Y. 2016.
\newblock Unitary evolution recurrent neural networks.
\newblock In \emph{Proceedings of The 33rd International Conference on Machine
  Learning}, volume~48, 1120--1128. PMLR.

\bibitem[{Bai, Kolter, and Koltun(2018)}]{TCN}
Bai, S.; Kolter, J.~Z.; and Koltun, V. 2018.
\newblock An empirical evaluation of generic convolutional and recurrent
  networks for sequence modeling.
\newblock \emph{arXiv preprint arXiv:1803.01271}.

\bibitem[{Bellec et~al.(2018)Bellec, Salaj, Subramoney, Legenstein, and
  Maass}]{LSNN}
Bellec, G.; Salaj, D.; Subramoney, A.; Legenstein, R.; and Maass, W. 2018.
\newblock Long Short-Term Memory and Learning-to-Learn in Networks of Spiking
  Neurons.
\newblock In \emph{Proceedings of the 32nd International Conference on Neural
  Information Processing Systems}, 795–805.

\bibitem[{Davies et~al.(2018)Davies, Srinivasa, Lin, Chinya, Cao, Choday,
  Dimou, Joshi, Imam, Jain et~al.}]{davies2018loihi}
Davies, M.; Srinivasa, N.; Lin, T.-H.; Chinya, G.; Cao, Y.; Choday, S.~H.;
  Dimou, G.; Joshi, P.; Imam, N.; Jain, S.; et~al. 2018.
\newblock Loihi: A neuromorphic manycore processor with on-chip learning.
\newblock \emph{Ieee Micro}, 38(1): 82--99.

\bibitem[{Devlin et~al.(2018)Devlin, Chang, Lee, and Toutanova}]{2018bert}
Devlin, J.; Chang, M.-W.; Lee, K.; and Toutanova, K. 2018.
\newblock Bert: Pre-training of deep bidirectional transformers for language
  understanding.
\newblock \emph{arXiv preprint arXiv:1810.04805}.

\bibitem[{DeWolf(2021)}]{dewolf2021spiking}
DeWolf, T. 2021.
\newblock Spiking neural networks take control.
\newblock \emph{Science Robotics}, 6(58): eabk3268.

\bibitem[{Ding et~al.(2022)Ding, Dong, Heide, Ding, Zhou, Yin, and
  Yang}]{BDETT}
Ding, J.; Dong, B.; Heide, F.; Ding, Y.; Zhou, Y.; Yin, B.; and Yang, X. 2022.
\newblock Biologically Inspired Dynamic Thresholds for Spiking Neural Networks.
\newblock In \emph{Advances in Neural Information Processing Systems}.

\bibitem[{Fang et~al.(2021{\natexlab{a}})Fang, Yu, Chen, Huang, Masquelier, and
  Tian}]{fang2021deep}
Fang, W.; Yu, Z.; Chen, Y.; Huang, T.; Masquelier, T.; and Tian, Y.
  2021{\natexlab{a}}.
\newblock Deep residual learning in spiking neural networks.
\newblock \emph{Advances in Neural Information Processing Systems}, 34:
  21056--21069.

\bibitem[{Fang et~al.(2021{\natexlab{b}})Fang, Yu, Chen, Masquelier, Huang, and
  Tian}]{PLIF}
Fang, W.; Yu, Z.; Chen, Y.; Masquelier, T.; Huang, T.; and Tian, Y.
  2021{\natexlab{b}}.
\newblock Incorporating learnable membrane time constant to enhance learning of
  spiking neural networks.
\newblock In \emph{Proceedings of the IEEE/CVF International Conference on
  Computer Vision}, 2661--2671.

\bibitem[{Fontaine, Pe{\~n}a, and Brette(2014)}]{fontaine2014spike}
Fontaine, B.; Pe{\~n}a, J.~L.; and Brette, R. 2014.
\newblock Spike-threshold adaptation predicted by membrane potential dynamics
  in vivo.
\newblock \emph{PLoS computational biology}, 10(4): e1003560.

\bibitem[{Gehring et~al.(2017)Gehring, Auli, Grangier, Yarats, and
  Dauphin}]{gehring2017convolutional}
Gehring, J.; Auli, M.; Grangier, D.; Yarats, D.; and Dauphin, Y.~N. 2017.
\newblock Convolutional sequence to sequence learning.
\newblock In \emph{Proceedings of the 34th International Conference on Machine
  Learning}, volume~70, 1243--1252. PMLR.

\bibitem[{G{\"u}tig(2016)}]{gutig2016spiking}
G{\"u}tig, R. 2016.
\newblock Spiking neurons can discover predictive features by aggregate-label
  learning.
\newblock \emph{Science}, 351(6277): aab4113.

\bibitem[{Hochreiter and Schmidhuber(1997)}]{LSTM97}
Hochreiter, S.; and Schmidhuber, J. 1997.
\newblock Long short-term memory.
\newblock \emph{Neural computation}, 9(8): 1735--1780.

\bibitem[{Horowitz(2014)}]{CMOS}
Horowitz, M. 2014.
\newblock 1.1 computing's energy problem (and what we can do about it).
\newblock In \emph{2014 IEEE International Solid-State Circuits Conference
  Digest of Technical Papers (ISSCC)}, 10--14. IEEE.

\bibitem[{Izhikevich(2004)}]{neurons}
Izhikevich, E.~M. 2004.
\newblock Which model to use for cortical spiking neurons?
\newblock \emph{IEEE transactions on neural networks}, 15(5): 1063--1070.

\bibitem[{Le, Jaitly, and Hinton(2015)}]{smnist}
Le, Q.~V.; Jaitly, N.; and Hinton, G.~E. 2015.
\newblock A simple way to initialize recurrent networks of rectified linear
  units.
\newblock \emph{arXiv preprint arXiv:1504.00941}.

\bibitem[{Li et~al.(2018)Li, Li, Cook, Zhu, and Gao}]{indrnn}
Li, S.; Li, W.; Cook, C.; Zhu, C.; and Gao, Y. 2018.
\newblock Independently recurrent neural network (indrnn): Building a longer
  and deeper rnn.
\newblock In \emph{Proceedings of the IEEE conference on computer vision and
  pattern recognition}, 5457--5466.

\bibitem[{Merolla et~al.(2014)Merolla, Arthur, Alvarez-Icaza, Cassidy, Sawada,
  Akopyan, Jackson, Imam, Guo, Nakamura et~al.}]{merolla2014million}
Merolla, P.~A.; Arthur, J.~V.; Alvarez-Icaza, R.; Cassidy, A.~S.; Sawada, J.;
  Akopyan, F.; Jackson, B.~L.; Imam, N.; Guo, C.; Nakamura, Y.; et~al. 2014.
\newblock A million spiking-neuron integrated circuit with a scalable
  communication network and interface.
\newblock \emph{Science}, 345(6197): 668--673.

\bibitem[{Neftci, Mostafa, and Zenke(2019)}]{8891809}
Neftci, E.~O.; Mostafa, H.; and Zenke, F. 2019.
\newblock Surrogate Gradient Learning in Spiking Neural Networks: Bringing the
  Power of Gradient-Based Optimization to Spiking Neural Networks.
\newblock \emph{IEEE Signal Processing Magazine}, 36(6): 51--63.

\bibitem[{Pei et~al.(2019)Pei, Deng, Song, Zhao, Zhang, Wu, Wang, Zou, Wu, He
  et~al.}]{pei2019towards}
Pei, J.; Deng, L.; Song, S.; Zhao, M.; Zhang, Y.; Wu, S.; Wang, G.; Zou, Z.;
  Wu, Z.; He, W.; et~al. 2019.
\newblock Towards artificial general intelligence with hybrid Tianjic chip
  architecture.
\newblock \emph{Nature}, 572(7767): 106--111.

\bibitem[{Qin et~al.(2023)Qin, Wang, Yan, and Tang}]{qin2023attention}
Qin, L.; Wang, Z.; Yan, R.; and Tang, H. 2023.
\newblock Attention-Based Deep Spiking Neural Networks for Temporal Credit
  Assignment Problems.
\newblock \emph{IEEE Transactions on Neural Networks and Learning Systems},
  1--11.

\bibitem[{Radford et~al.(2019)Radford, Wu, Child, Luan, Amodei, Sutskever
  et~al.}]{GPT2}
Radford, A.; Wu, J.; Child, R.; Luan, D.; Amodei, D.; Sutskever, I.; et~al.
  2019.
\newblock Language models are unsupervised multitask learners.
\newblock \emph{OpenAI blog}, 1(8): 9.

\bibitem[{Rao et~al.(2022)Rao, Plank, Wild, and Maass}]{AHP}
Rao, A.; Plank, P.; Wild, A.; and Maass, W. 2022.
\newblock A long short-term memory for AI applications in spike-based
  neuromorphic hardware.
\newblock \emph{Nature Machine Intelligence}, 4(5): 467--479.

\bibitem[{Shaban, Bezugam, and Suri(2021)}]{DEXAT}
Shaban, A.; Bezugam, S.~S.; and Suri, M. 2021.
\newblock An adaptive threshold neuron for recurrent spiking neural networks
  with nanodevice hardware implementation.
\newblock \emph{Nature Communications}, 12(1): 4234.

\bibitem[{Sunkin et~al.(2012)Sunkin, Ng, Lau, Dolbeare, Gilbert, Thompson,
  Hawrylycz, and Dang}]{sunkin2012allen}
Sunkin, S.~M.; Ng, L.; Lau, C.; Dolbeare, T.; Gilbert, T.~L.; Thompson, C.~L.;
  Hawrylycz, M.; and Dang, C. 2012.
\newblock Allen Brain Atlas: an integrated spatio-temporal portal for exploring
  the central nervous system.
\newblock \emph{Nucleic acids research}, 41(D1): D996--D1008.

\bibitem[{Vaswani et~al.(2017)Vaswani, Shazeer, Parmar, Uszkoreit, Jones,
  Gomez, Kaiser, and Polosukhin}]{vaswani2017attention}
Vaswani, A.; Shazeer, N.; Parmar, N.; Uszkoreit, J.; Jones, L.; Gomez, A.~N.;
  Kaiser, {\L}.; and Polosukhin, I. 2017.
\newblock Attention is all you need.
\newblock \emph{Advances in neural information processing systems}, 30.

\bibitem[{Wang, Cheng, and Lim(2022)}]{LTMD}
Wang, S.; Cheng, T.~H.; and Lim, M.-H. 2022.
\newblock LTMD: Learning Improvement of Spiking Neural Networks with Learnable
  Thresholding Neurons and Moderate Dropout.
\newblock \emph{Advances in Neural Information Processing Systems}, 35:
  28350--28362.

\bibitem[{Wang et~al.(2023)Wang, Jiang, Lian, Yan, and Tang}]{wang2023adaptive}
Wang, Z.; Jiang, R.; Lian, S.; Yan, R.; and Tang, H. 2023.
\newblock Adaptive Smoothing Gradient Learning for Spiking Neural Networks.
\newblock In \emph{Proceedings of the 40th International Conference on Machine
  Learning}, volume 202, 35798--35816. PMLR.

\bibitem[{Wu et~al.(2021{\natexlab{a}})Wu, Chua, Zhang, Li, Li, and
  Tan}]{wu2021tandem}
Wu, J.; Chua, Y.; Zhang, M.; Li, G.; Li, H.; and Tan, K.~C. 2021{\natexlab{a}}.
\newblock A Tandem Learning Rule for Effective Training and Rapid Inference of
  Deep Spiking Neural Networks.
\newblock \emph{IEEE Transactions on Neural Networks and Learning Systems},
  1--15.

\bibitem[{Wu et~al.(2018{\natexlab{a}})Wu, Chua, Zhang, Li, and
  Tan}]{wu2018spiking}
Wu, J.; Chua, Y.; Zhang, M.; Li, H.; and Tan, K.~C. 2018{\natexlab{a}}.
\newblock A spiking neural network framework for robust sound classification.
\newblock \emph{Frontiers in neuroscience}, 12: 836.

\bibitem[{Wu et~al.(2021{\natexlab{b}})Wu, Xu, Han, Zhou, Zhang, Li, and
  Tan}]{wu2021progressive}
Wu, J.; Xu, C.; Han, X.; Zhou, D.; Zhang, M.; Li, H.; and Tan, K.~C.
  2021{\natexlab{b}}.
\newblock Progressive tandem learning for pattern recognition with deep spiking
  neural networks.
\newblock \emph{IEEE Transactions on Pattern Analysis and Machine
  Intelligence}, 1--1.

\bibitem[{Wu et~al.(2020)Wu, Y{\i}lmaz, Zhang, Li, and Tan}]{wu2020deep}
Wu, J.; Y{\i}lmaz, E.; Zhang, M.; Li, H.; and Tan, K.~C. 2020.
\newblock Deep spiking neural networks for large vocabulary automatic speech
  recognition.
\newblock \emph{Frontiers in neuroscience}, 14: 199.

\bibitem[{Wu et~al.(2018{\natexlab{b}})Wu, Deng, Li, Zhu, and
  Shi}]{wu2018spatio}
Wu, Y.; Deng, L.; Li, G.; Zhu, J.; and Shi, L. 2018{\natexlab{b}}.
\newblock Spatio-temporal backpropagation for training high-performance spiking
  neural networks.
\newblock \emph{Frontiers in neuroscience}, 12: 331.

\bibitem[{Yao et~al.(2022{\natexlab{a}})Yao, Zhao, Zhang, Hu, Deng, Tian, Xu,
  and Li}]{TASNN}
Yao, M.; Zhao, G.; Zhang, H.; Hu, Y.; Deng, L.; Tian, Y.; Xu, B.; and Li, G.
  2022{\natexlab{a}}.
\newblock Attention spiking neural networks.
\newblock \emph{arXiv preprint arXiv:2209.13929}.

\bibitem[{Yao et~al.(2022{\natexlab{b}})Yao, Li, Mo, and Cheng}]{GLIF}
Yao, X.; Li, F.; Mo, Z.; and Cheng, J. 2022{\natexlab{b}}.
\newblock Glif: A unified gated leaky integrate-and-fire neuron for spiking
  neural networks.
\newblock \emph{Advances in Neural Information Processing Systems}, 35:
  32160--32171.

\bibitem[{Yin, Corradi, and Boht{\'e}(2021)}]{ALIF}
Yin, B.; Corradi, F.; and Boht{\'e}, S.~M. 2021.
\newblock Accurate and efficient time-domain classification with adaptive
  spiking recurrent neural networks.
\newblock \emph{Nature Machine Intelligence}, 3(10): 905--913.

\bibitem[{Zhang and Linden(2003)}]{zhang2003other}
Zhang, W.; and Linden, D.~J. 2003.
\newblock The other side of the engram: experience-driven changes in neuronal
  intrinsic excitability.
\newblock \emph{Nature Reviews Neuroscience}, 4(11): 885--900.

\bibitem[{Zheng et~al.(2021)Zheng, Wu, Deng, Hu, and Li}]{zheng2021going}
Zheng, H.; Wu, Y.; Deng, L.; Hu, Y.; and Li, G. 2021.
\newblock Going deeper with directly-trained larger spiking neural networks.
\newblock In \emph{Proceedings of the AAAI conference on artificial
  intelligence}, volume~35, 11062--11070.

\end{thebibliography}

\onecolumn
\pagestyle{plain}
\clearpage
\newpage
\setcounter{page}{1}

\Large
\begin{center}
    {\bf \LARGE Supplementary Materials}\\
    \vspace{0.2cm}
    { \Large Unleashing the Potential of Spiking Neural Networks for \\ Sequential Modeling with Contextual Embedding}
\end{center}
\rule[-0.5pt]{18.1cm}{0.06em}

\parskip=2pt

\large

\parskip=2pt

\noindent

\subsection{\Large Gradient Descent of the ALIF Model}
As mentioned in the main text, the dynamic of the ALIF model could be formulated as follows: 

\begin{equation}
\left\{\begin{array}{lr}
\Theta[t]=\beta \cdot (\Theta[t-1]-\Theta_0) + \gamma \cdot S[t-1]+\Theta_0,&  \\
V[t]=\alpha \cdot V[t-1] \cdot (1-S[t-1])+I[t],& \\
S[t]=H\left(V[t]-\Theta[t]\right), &
\end{array}\right.
\end{equation}

Based on that, we could derive the gradient propagation of the ALIF model as follows:
\begin{equation}
\begin{aligned}
\frac{\partial \mathcal{L} }{\partial V^l[t]}&=\frac{\partial \mathcal{L} }{\partial S^l[t]}\frac{\partial S^l[t]}{\partial V^l[t]}+\frac{\partial \mathcal{L} }{\partial V^l[t+1]}\frac{\partial V^l[t+1]}{\partial V^l[t]} \\
&=\frac{\partial \mathcal{L} }{\partial S^l[t]} H^{\prime}(V^l[t]-\Theta^l[t])+\frac{\partial \mathcal{L} }{\partial V^l[t+1]} \alpha(1-S^l[t]),\\
 \frac{\partial \mathcal{L} }{\partial \Theta^l[t]}&=\frac{\partial \mathcal{L} }{\partial S^l[t]} \frac{\partial S^l[t]}{\partial \Theta^l[t]}+\frac{\partial \mathcal{L} }{\partial \Theta^l[t+1]} \frac{\partial \Theta^l[t+1]}{\partial \Theta^l[t]} \qquad \qquad \qquad \qquad \qquad \qquad\\
 &=-\frac{\partial \mathcal{L} }{\partial S^l[t]} H^{\prime}(V^l[t]-\Theta^l[t])+\frac{\partial \mathcal{L} }{\partial \Theta^l[t+1]} \beta \\  
 &=\sum_{i=t}^{T}{ {-\frac{\partial \mathcal{L} }{\partial S^l[i]}H^{\prime}(V^l[i]-\Theta^l[i])\beta^{i-t}}},\\
\frac{\partial \mathcal{L} }{\partial S^l[t]}&=\frac{\partial \mathcal{L} }{\partial V^l[t+1]} \frac{\partial V^l[t+1]}{\partial S^l[t]}+\frac{\partial \mathcal{L} }{\partial V^{l+1}[t]} \frac{\partial V^{l+1}[t]}{\partial S^l[t]}+ \frac{\partial \mathcal{L} }{\partial \Theta^l[t+1]}\frac{\partial \Theta^l[t+1]}{\partial S^l[t]} \\
&=-\frac{\partial \mathcal{L} }{\partial V^l[t+1]}\alpha V^l[t]+\frac{\partial \mathcal{L} }{\partial V^{l+1}[t]} W^{l} + \frac{\partial \mathcal{L} }{\partial \Theta^l[t+1]}\gamma
\end{aligned}
\end{equation}

By Substituting $\frac{\partial \mathcal{L} }{\partial \Theta[t+1]}$ into $\frac{\partial \mathcal{L} }{\partial S[t]}$, we can obtain:

\begin{equation}
\label{eq:alif_grad}
    \frac{\partial \mathcal{L} }{\partial S[t]}=\underbrace{-\frac{\partial \mathcal{L} }{\partial V[t+1]}\alpha V[t]}_{\text{LIF (Temporal)}}+\underbrace{\frac{\partial \mathcal{L} }{\partial V^{l+1}[t]} W^{l, l+1}}_{\text {LIF (Spatial)}} -\underbrace{\gamma\sum_{i=t+1}^{T}{ {\frac{\partial \mathcal{L} }{\partial S[i]}H^{\prime}(V[i]-\Theta[i])\beta^{i-t-1}}}}_{\text{Adaptive threshold}}
\end{equation}

As can be seen from Eq. \ref{eq:alif_grad}, the main contribution of adaptive threshold to temporal credit assignment lies in the last term. Benefiting from the slow decay and non-reset characteristic, the ALIF model could reserve memory longer than the LIF model. However, ALIF still suffers from two crucial limitations that prevent the backpropagation of gradients to very long spans, which leads to its failure to resolve long-term sequential modeling problems. Firstly, due to the involvement of surrogate gradient $H^{\prime}(V[i]-\Theta[i])$ in the last term, the temporal dependency between timestep $t$ with timestep $i$ will be eliminated if $|V^l[t]-\Theta^l[t]|\geq \Gamma$. Such insufficient temporal credit assignment significantly hinders the establishment of the integral temporal structure.
Secondly, since the decay of threshold $\gamma$ is a constant, the backpropagated gradient from the subsequent timestep $t$ to the previous timesteps will decay proportionally with $\beta$ as the time span increases. Specifically, the gradient received by timestep $t+1$ is always greater than timestep $t$. Therefore, neurons are unable to establish more adequate memories based on the importance of the current timestep to forthcoming timesteps. On the contrary, our proposed CE-LIF model could address the aforementioned problems by introducing the interaction between temporal encoding and membrane threshold in threshold adaptation to enhance memory storage. The detailed analysis is provided in the Subsection \textbf{Impact of TE on Gradient Propagation} of the main text.

\subsection{\Large Comparison of CE-LIF within Feedforward and Recurrent Structures}
The results in Table \ref{tab:mnist} indicate the fact that recurrent structure always outperforms feedforward structure when they are incorporated with the same spiking neuron model. However, the SOTA results of our CE-LIF model are achieved with a feedforward network structure. Then the question arises: how does CE-LIF perform within a recurrent structure compared with the feedforward structure? Therefore, we further conduct an extensive study to gain insight into the impact of recurrent structures on the CE-LIF model.

\begin{table}[h]
    \caption{Performance of CE-LIF within feedforward and recurrent structures on pixel-level sequential image classification tasks.}
\label{tab:fffb}
\begin{center}
\resizebox{0.7\textwidth}{!}{
\begin{tabular}{lcccccc}
\hline
\multicolumn{1}{c}{\bf Model} &\multicolumn{1}{c}{\bf Architecture} &\multicolumn{1}{c}{\bf Parameters}  &\multicolumn{1}{c}{\bf Hidden dimensions}  &\multicolumn{1}{c}{\bf Seq-MNIST}  &\multicolumn{1}{c}{\bf P-MNIST} 
\\ 
\hline
\multirow{3}{*}{CE-LIF} & Feedforward & 153.9k & [64,152,152] & 98.84\% &  98.14\%\\
                        & Recurrent  & 158.3k & [64, 136, 136] & 98.41\% & 97.83\% \\
                        & Recurrent & 181.4k & [64, 152, 152] & 98.91\% & 98.09\% \\
\hline
\end{tabular}}
\end{center}
\end{table}

Given that the introduction of recurrent connections leads to an increment of parameter counts, Table \ref{tab:fffb} presents the performance of the CE-LIF neuron within recurrent structures in two scenarios. The first one employs the same hidden dimensions as the feedforward structure but accommodates larger parameter counts, while the other one retains the same parameter count as the feedforward structure but necessitates a reduction in network scale. The results indicate that the inclusion of recurrent connections does not notably bolster the model's performance when the architectural dimension remains the same but increases the network parameters by 18\%. Such observation indicates that the temporal credit assignment in CE-LIF neurons within the feedforward structure is enough to extract the sequential order information, and the network performance has already reached saturation. Therefore, it is not necessary to incorporate more feedback signals via recurrent connections to construct more complex dynamics. On the other hand, with the reduction of network scale, the recurrent structure even performs worse than feedforward ones under roughly the same parameter counts, which further supports our justification that feedforward structure is the better choice for the implementation of CE-LIF neurons in sequential modeling.

\subsection{\Large Computational Efficiency of CE-LIF}
One of the most impressive advantages of the CE-LIF neuron is it can achieve superior sequential modeling performance with purely feedforward connections. Such concise structure not only contributes to stable convergence but also results in high energy efficiency.  

Therefore, we will provide further elucidation about the computational cost of CE-LIF, LIF, ALIF, and LSTM models. Theoretically, we utilize the number of Multiply-Accumulate (MAC) operations and cheap Accumulate (AC) operations to quantify the energy consumption during one inference. As an artificial neural network (ANN), LSTM induces MACs in hidden state updating and information transferring between layers.
Benefiting from the sparse spike trains, ALIF could use more energy-efficient AC operations to propagate information between neurons. However, computing recurrent signals still introduces $n \times n $ ACs in each timestep, which significantly increases total computational cost. Our CE-LIF spiking model adopts feedforward connection only, eliminating the cost associated with recurrent connections. 
Despite the fact that the update of the threshold in CE-LIF requires extra operations, such computation is element-wise, resulting in substantial energy savings totally.

 Following the data collected on the $45nm$ CMOS \cite{CMOS}, we set $E_{AC} = 0.9\ pJ$ and $E_{MAC} = 4.6\ pJ$ respectively. Additionally, in order to get the spiking sparsity of neurons, we perform inference on all spiking models with a randomly selected input batch and record the average spike frequency of each presynaptic and postsynaptic layers, namely, $Fr_{in}$ and $Fr_{out}$, these frequencies are subsequently used to estimate the number of AC operations. The recorded spike frequencies of all hidden layers in the LIF model and CE-LIF model are $[0.22, 0.145, 0.004]$ and $[0.22,0.23,0.44]$, respectively. The frequency of ALIF is computed based on their reported results \cite{ALIF}.

For a fair comparison, all models adopt the same parameters about $155k$, consisting of three hidden layers and one classification layer. The theoretical and empirical results are both presented in Table \ref{tab:mac}. The results clearly indicate that the feedforward network with CE-LIF neurons could achieve higher energy efficiency, spacing more than 50 $\times$ computational cost compared with LSTM.

\begin{table}[h]
\caption{Comparative analysis of energy cost among diverse models with equivalent parameter counts}
\label{tab:mac}
\begin{center}
\resizebox{0.8\textwidth}{!}{
\begin{tabular}{lccccc}
\multicolumn{1}{c}{\bf Neuron Model} &\multicolumn{1}{c}{\bf Architecture} &\multicolumn{1}{c}{\bf Theoretical Energy Cost}  &\multicolumn{1}{c}{\bf Empirical Energy Costs}
\\ \hline 
LSTM & RNN & \thead{$4(mn+nn)E_{MAC}$\\$+ 17nE_{MAC}$}  & 774.9 nJ\\ 
\hline

LIF  &SRNN   & \thead{$mn F r_{in} E_{A C}+(nn+n)Fr_{out} E_{AC}$\\$+n E_{M A C}$} & 24.1 nJ\\
\hline
ALIF    &SRNN   & \thead{$mnFr_{in}E_{AC}+(nn+2n)Fr_{out}E_{AC}$ \\$+4nE_{MAC}$} & 22.9 nJ\\ 
\hline
\textbf{CE-LIF(Ours)}     &\textbf{FFSNN}   & \thead{$(mnFr_{in}+nFr_{out})E_{AC}$\\$+3nE_{MAC}$} & \textbf{13.5 nJ}\\
\hline

\end{tabular}}
\end{center}
\end{table}

\subsection{\Large Loss landscapes}
\begin{figure*}[h]
\centering
\subfigure[]{
	\includegraphics[trim=30 0 30 0,clip,scale=0.35]{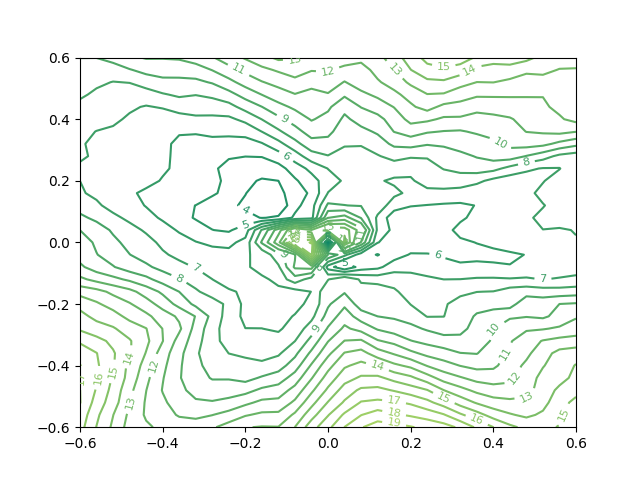}}\hspace{0em}
  \subfigure[]{
	\includegraphics[trim=30 0 30 0,clip,scale=0.35]{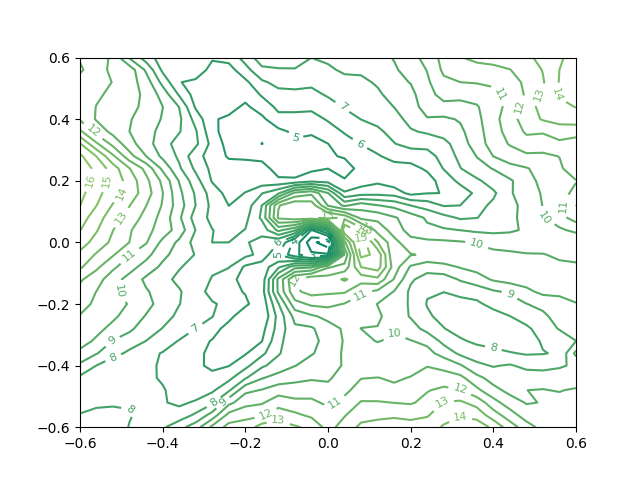}}\hspace{0em}
 \subfigure[]{
	\includegraphics[trim=30 0 30 0,clip,scale=0.35]{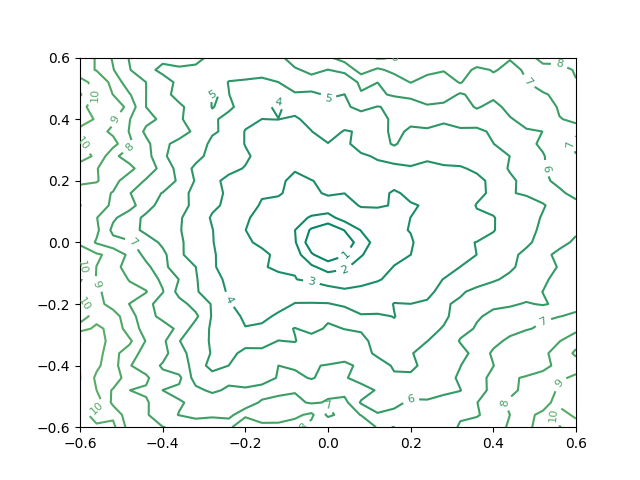}}\hspace{0em}
\caption{ Two-dimensional contour plots of loss landscapes of \textbf{(a)} LIF, \textbf{(b)} ALIF, and \textbf{(c)} CE-LIF models. LIF and ALIF models adopt the recurrent structure while the CE-LIF model implements the feedforward structure.}
\label{fig: loss_s}
\vspace{-4mm}
\end{figure*}

In Figure \ref{fig: loss}, we demonstrate the 3D plots of loss landscapes of LIF, ALIF, and CE-LIF models. However, such 3D plots lack a detailed depiction of the smoothness of the loss close to the origin. Therefore, we also compared the 2D contour plots of the loss landscape as presented in Figure \ref{fig: loss_s}. Compared with the other two recurrent models, our CE-LIF model exhibits smoother contours with clearer and evener downward trends, indicating the convergence of CE-LIF is much easier than other models regardless of initialization or training settings.

\subsection{\Large Experimental Setups}
In this section, we present the detailed experimental setups, including datasets, model architectures, and specific hyperparameters used in each task.
\subsection{Datasets}
Our experiments are conducted on Seq-MNIST, PS-MNIST, Adding Problem, and Copy Memory datasets. 
\begin{itemize}
    \item \textbf{Seq-MNIST and PS-MNIST} \cite{smnist}
    are two datasets developed from MNIST handwritten digit recognition dataset. In these tasks, the $28 \times 28$ gray-scaled image is read into the network through a raster scan, one pixel at a time. In the Seq-MNIST task, the sequential order is identified to how humans read: row-by-row. On the other hand, in the PS-MNIST task, the order is shuffled by a fixed random permutation matrix before processing by the network, significantly enhancing the task difficulty. 
    \item \textbf{Adding Problem} \cite{LSTM97, URNN} is a regression task that aims to generate solutions that require adding two numbers based on two specific sequences. The first value sequence contains $T$ floats sampled from $(0,1)$ randomly, while the second indicator sequence assigns a value of $1$ to two randomly selected entries and $0$ to the remaining entries. The desired output of the network is the sum of two numbers from the first sequence, whose locations are where $1$ appears in the second sequence.
    \item \textbf{Copy Memory} \cite{LSTM97,URNN}. A synthetic classification task that involves remembering and retrieving a sequence of numbers. The network should learn to remember the digits sequence seen T timesteps earlier in this task and output it sequentially in the output phase.
\end{itemize}

\subsection{Model Architecture and Hyperparameters}
All experiments in this work adopt a uniform network architecture comprising three hidden layers and one classification layer. \textbf{It should be noted that each layer of the network shares the same set of TE vectors.} Here, we specify all task-specific hyperparameters and architecture configurations as follows:
\begin{table}[h]
\normalsize
\centering
\caption{ Hyper-parameters used for reported results
}
\label{tab:hyperpara}
\resizebox{1\textwidth}{!}{%
\begin{tabular}{l c c c c c c c c}
\hline
\hline
\textbf{Datasets}  & \textbf{Hidden Dimensions} &\textbf{Learning Rate} &\textbf{Batch Size} &\textbf{TE Initialization}  &\textbf{$\alpha$}  &\textbf{$\beta$} &\textbf{$\Theta_0$} &\textbf{$\Gamma$}  \\ 
\hline
Seq-MNIST        &[64,88,88]/[64,152,152] &5e-4 &256  &$\mathcal{N}(0.01,0.01)$   &0.5 &0.99  &0.3  &0.2  \\
PS-MNIST        &[64,88,88]/[64,152,152] &5e-4 &256   &$\mathcal{N}(0.01,0.01)$   &0.5 &0.99  &0.3 &0.2 \\
Adding Problem   &[64,256,256] &5e-4 &256 &$\mathcal{N}(0.01,0.01)$   &0.5 &0.99  &0.3 &0.2 \\
Copy Memory       &[64,256,256] &1e-3 &256  &$\mathcal{N}(0.01,0.01)$   &0.5 & $1-1/T$  &0.3  &0.2  \\

\hline
\hline
\end{tabular}}
\end{table}

\subsection{Experiment Specific Configuration}
\begin{itemize}
    \item \textbf{Illustration of the evolution of neuronal variables during the Seq-MNIST classification task (Figure \ref{fig: out_ana}):} In this experiment, we employ an output decoding strategy where the network output is defined as the average spike frequency of CE-LIF neurons across all timesteps. This strategy aims to help readers to get a more intuitive understanding of the temporal pattern formed by the CE-LIF model. Specifically, the threshold adaptation, spike bursting dynamics, and labels of the classification neurons could be visualized with the input images simultaneously. For other experiments, we adhere to the convention followed by other SNNs \cite{PLIF,wu2021tandem} to decode the network output by averaging the output of a linear classifier across all time steps.
    \item \textbf{Efficacy of CE in Facilitating Long-term TCA:} In this experiment, a set of $256$ training samples are taken as input for each model. The overall loss is determined by applying the cross-entropy (CE) loss function to the network outputs and corresponding labels. During the backpropagation process, the loss of each neuron in each timestep denoted as $Grad(n,t)$ is recorded. To ensure clarity, the gradient value is normalized as $Grad(t,n)=\frac{Grad(t,n)}{\sum^N_{i=0}\sum^T_{j=0}Grad(i,j)}$ for visualization purposes. In order to enhance the consistency, the $Grad(t,:)$ mentioned here is denoted as $Grad[t]$ in the main text.

\end{itemize}

\subsection{\Large Computing Infrastructure}
All experiments are performed on a computer equipped with GeForce GTX 3090Ti GPUs (24G Memory) and Pytorch 1.13.0.

\subsection{\Large Code Availability}
The source codes used in this work are provided in a zip file and will be publicly available after the review process.

\end{document}